# Deep Dreams Are Made of This: Visualizing Monosemantic Features in Diffusion Models

**Adam Szokalski**
Institute of Computer Science, Warsaw University of Technology, Warsaw, Poland

**Mateusz Modrzejewski**
Institute of Computer Science, Warsaw University of Technology, Warsaw, Poland

## Abstract

This paper proposes latent visualization by optimization (LVO), a mechanistic interpretability technique that extends feature visualization by optimization—originally developed for convolutional neural networks—to latent diffusion models. LVO employs sparse autoencoders (SAEs) to disentangle polysemantic layer representations into monosemantic features. Key contributions include latent-space optimization, time-step activity analysis, schedule-matched noise injection, prior initialization through feature steering, and suitable regularization strategies. We demonstrate the method on Stable Diffusion 1.5 fine-tuned on the Style50 dataset, showing that SAE features produce clear visualizations of recognizable concepts—including diagonal compositions, human figures, roses, cables, and waterfall foam—that correlate with dataset examples, while the baseline without disentanglement produces less coherent results. We further show that regularization techniques from pixel-space feature visualization transfer to the latent domain, though they require different configurations for the raw-layer and SAE variants. Compared to dataset examples and steering, LVO provides complementary insights by directly revealing what activates a feature rather than its downstream effects.

## 1 Introduction

Deep neural networks have become the state-of-the-art approach to many modern AI tasks and are often deployed in high-stakes scenarios. However, their inner workings remain largely mysterious, essentially making them black boxes. These concerns motivate mechanistic interpretability, a field that aims to reverse-engineer neural networks to understand their internal mechanisms [1], [2].

One of the key techniques in mechanistic interpretability of image models is feature visualization, which generates synthetic inputs that maximally activate specific internal features [3], [4], [5]. Sparse autoencoders (SAEs) address a fundamental issue in this setting: raw neural features are often polysemantic and represent multiple unrelated concepts simultaneously, whereas SAEs can disentangle them into more fundamental monosemantic features [6], [7].

Latent diffusion models (LDMs) have become the standard architecture for image generation, yet their iterative denoising process and operation in a compressed variational-autoencoder (VAE) latent space make classical feature-visualization methods inapplicable without modification [8], [9]. Existing diffusion-model interpretability methods, such as dataset examples and steering, provide valuable insights but each has limitations. Dataset examples are constrained by the dataset's content and may not capture the full range of a feature's functionality [4]. Steering, in turn, shows downstream effects of amplifying a feature rather than the cause of its activation.

Building on recent progress in diffusion-model SAEs [7], we propose **latent visualization by optimization** (LVO), a feature-visualization-by-optimization technique designed for LDMs that reveals what individual features have learned to represent. To evaluate the effect of disentanglement, we compare the SAE-based instantiation of LVO with a baseline that optimizes raw layer activations directly.

**Research questions** We investigate four questions: **RQ1**. Can feature visualization by optimization be extended to latent diffusion models to produce human-interpretable results? **RQ2**. Does SAE disentanglement improve interpretability relative to optimizing raw layer activations? **RQ3**. How do regularization techniques from pixel-space feature visualization transfer to the latent domain? **RQ4**. How do the resulting visualizations compare qualitatively with dataset examples and steering?

**Contributions** This paper makes four contributions. First, we **propose LVO**, which extends feature visualization by optimization to latent diffusion models through time-step activity analysis, latent-space optimization, schedule-matched noise injection, prior initialization via steering, and latent-specific regularization. Second, we provide an explicit raw-layer baseline that isolates the effect of SAE disentanglement. Third, we report the first systematic ablation of pixel-space feature-visualization regularizers in the LDM latent domain, showing that the same regularizers behave differently with and without disentanglement, and we introduce two new latent-specific penalties. Fourth, we demonstrate that LVO complements dataset examples and steering by directly revealing what activates a feature.

## 2 Method

We propose **latent visualization by optimization** (LVO), an extension of feature visualization by optimization [4] to the LDM setting. The core idea, inherited from the classical technique, is to optimize an input by gradient ascent so that it maximally activates a target feature, producing an image that represents the ideal input for exciting the feature. LVO introduces four diffusion-specific components on top of this core: time-step activity analysis, latent-space optimization with schedule-matched noise, prior initialization through steering, and a regularization stack tailored to the latent domain. The overall pipeline is shown in Figure 1.

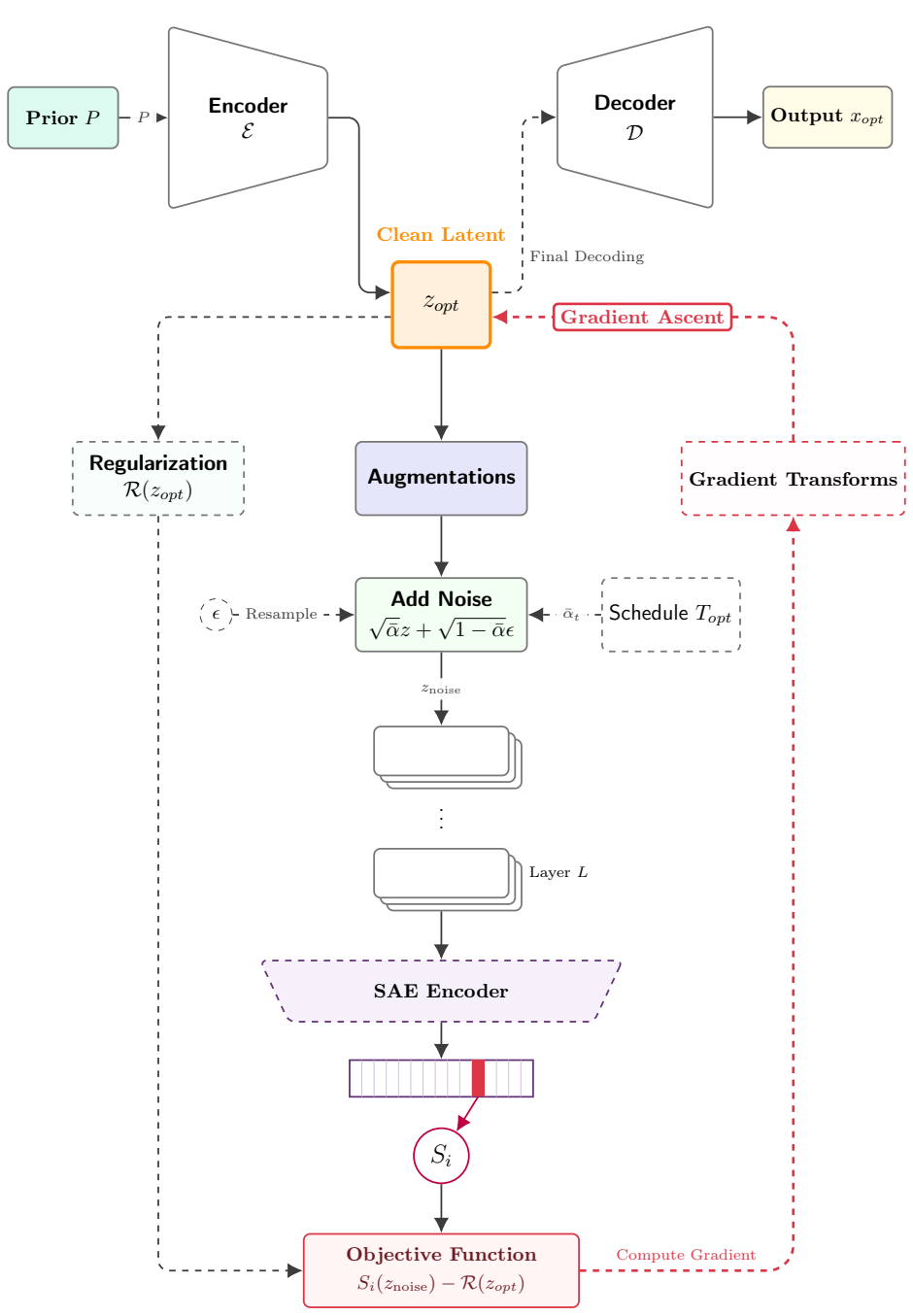

Figure 1: Method overview. The optimization is performed on a selected time-step. The clean latent $z_{\text{opt}}$ is initialized with the encoded prior $\mathcal{E}(P)$, where $P$ is a feature-steered image, and updated by gradient ascent on the target activation. After enough steps, $z_{\text{opt}}$ is decoded into the image $x_{\text{opt}}$ shown to the analyst.

### 2.1 Time-step activity analysis

LDMs denoise an image over a sequence of time-steps, and the same feature can be active at different time-steps for different reasons. A feature must therefore be interpreted under the conditions in which it is most active. Analysing all time-steps would require 1000 visualizations per feature for Stable Diffusion 1.X and would make the explanation nearly as complex as the model itself. We therefore restrict analysis to **activity peaks**.

To find them, we run the model on its training set and record the top-$k$ activating features at each time-step. For each feature and time-step, **activity** is defined as the frequency with which the feature appears among those top-$k$ activations across the dataset. This yields a time-step activity profile for every feature. We then select up to $p$ highest peaks using `SciPy`'s `find_peaks` with a minimum-separation constraint of 100 time-steps. In our experiments we use $k = 20$ and $p = 3$.

Each peak is treated as an independent visualization condition: the forward pass is conditioned on the corresponding time-step and a separate visualization is produced. Comparing across peaks reveals whether a feature plays a different role early or late in the denoising trajectory.

### 2.2 Latent-space optimization

Feature visualization by gradient ascent is well known to drift toward adversarial maxima dominated by high-frequency noise: patterns that maximize activation but do not look like anything to a human observer [3], [4]. The deeper the layer, the longer the gradient path, and A. Odena, V. Dumoulin, and C. Olah [10] show that each convolutional block contributes checkerboard patterns to that gradient. This is the central failure mode every regularizer in Section 2.6 addresses.

Optimization is performed in the VAE latent space rather than in pixel space. Treating the VAE as part of the model and back-propagating through it is roughly an order of magnitude slower in our setup, and the additional convolutional blocks of the decoder are themselves a major source of the checkerboard gradients above [9], [10].

Working in the latent introduces a second failure mode. In LDMs the VAE latent is typically scaled to a standard normal distribution, so most values fall in $[-3, 3]$ [9]. The decoder has rarely seen samples outside this range during training, and decoding them produces oversaturation and overexposure. We address this with two latent-specific penalties introduced in Section 2.6.

### 2.3 Schedule-matched noise

The U-Net inside an LDM is trained on inputs whose signal-to-noise ratio is fixed by the time-step that conditions denoising [8]. If the optimized latent is forwarded as-is at an intermediate time-step, the model sees a clean signal where it expects partial noise; the conditioning is broken and the visualization corresponds to a much later stage of the diffusion process.

To preserve the expected conditioning, we re-noise the latent on every forward pass using the scheduler coefficients of the target time-step:

$$z_{\text{noise}} = \sqrt{\bar{\alpha}_t} z + \sqrt{1 - \bar{\alpha}_t} \epsilon \tag{1}$$

where $z$ is the optimized latent, $t$ is the chosen time-step, and $\bar{\alpha}_t$ is the cumulative scheduler coefficient [8]. The sample $\epsilon \sim \mathcal{N}(0, I)$ is redrawn at every optimization step.

### 2.4 Prior initialization

Initializing optimization from a **prior image** keeps the result on the interpretable manifold by starting it in a region that already looks plausible [3], [4]. Hard parameterizations such as a generative-model prior [11] provide stronger guarantees, but soft DeepDream-style initialization is sufficient in our setting because the optimization landscape is highly non-convex and tends to settle in the nearest interpretable maximum [12].

We use the diffusion model itself to generate the prior, by **steering** the target feature at the analyzed layer. Adapting the SAeUron formulation [7], for an SAE feature $i$ at time-step $t$ we add a scaled decoder direction to the layer activations:

$$F_t \leftarrow F_t + \lambda_{i,t} \boldsymbol{d}_i, \quad \lambda_{i,t} = \gamma \cdot \max_{x \in \mathcal{D}} f_{i,t}(x) \tag{2}$$

where $\boldsymbol{d}_i$ is the $i$-th SAE decoder direction, $f_{i,t}(x)$ is the activation of feature $i$ on input $x$ at step $t$, $\mathcal{D}$ is the SAE training set, and $\gamma$ is a steering-strength parameter [13]. In the raw-layer baseline the same formula is applied directly to a single layer channel without the SAE encoding step.

To avoid biasing the prior toward time-steps where the feature is barely active, we steer only on time-steps with non-zero activity in the feature's profile. The resulting steered latent (denoted $P$ in Figure 1) is encoded once and used to initialize $z_{\text{opt}}$, so the encoder is bypassed during the optimization loop.

### 2.5 Target features and the raw-layer baseline

The target representation differs between the two LVO variants compared throughout this paper. The baseline optimizes a single raw activation channel of the analyzed layer. The SAE variant first encodes the layer with a sparse autoencoder trained on it [7] and optimizes a single feature in the SAE dictionary. The baseline therefore acts as a controlled comparison for the effect of disentanglement: any difference in interpretability between the two visualizations of the same layer is attributable to the SAE.

### 2.6 Regularization

The four techniques in this section attack the high-frequency-noise failure mode introduced in Section 2.2. Following the classical feature-visualization literature [4], [14], [15], we keep the search on the human-interpretable manifold with four pixel-space techniques. Two additional penalties are needed to handle the second, latent-domain failure mode (decoder out-of-range values) described in the same section.

**Transformation robustness** applies random jitter, rotation, and scaling to the latent before each forward pass [4]. Patterns that depend on a precise pixel arrangement are penalized; only features that remain salient under small geometric perturbations dominate the optimization.

**Spectral filtering** reshapes the gradient so that high-frequency components are attenuated [4]. Each step we take the FFT of the gradient, scale the spectrum by the inverse of the frequency, and return to the spatial domain. This biases the optimization trajectory toward smooth directions without altering the position of local maxima.

**Gradient smoothing** applies a Gaussian blur to the gradient with intensity decaying across iterations, so coarse structure is established first and fine details emerge only late in optimization [15].

**Total variation** [14] directly punishes neighbouring pixel differences. Although TV was originally defined in pixel space, modern VAEs preserve much of the input's 2D topology [9], so the same penalty remains effective when applied to the latent feature map:

$$R_{\text{TV}(x)} = \frac{1}{HW} \sum_{i,j} \left( |x_{i+1,j} - x_{i,j}| + |x_{i,j+1} - x_{i,j}| \right) \tag{3}$$

where $H$ and $W$ are its spatial dimensions.

**Range and moment penalties** address the latent-space failure mode. The **range penalty** uses an $L_2$ ReLU on values that leave the decoder's working range:

$$R_{\text{range}(z)} = \text{mean}\left(\text{ReLU}(|z| - 3)^2\right) \tag{4}$$

The **moment penalty** keeps the first two moments of the latent close to those of a standard normal through an $L_1$ penalty on the deviations:

$$R_{\text{moment}(z)} = |\text{mean}(z)| + \left|\sqrt{\text{var}(z)} - 1\right| \tag{5}$$

Both terms enter the objective as part of the regularization $\mathcal{R}(z_{\text{opt}})$ in Figure 1.

## 3 Experimental Setup

All experiments target the up-sampling attention layer `up_blocks.1.attentions.1` (`up1.1`) of Stable Diffusion 1.5 fine-tuned on the UnlearnCanvas Style50 dataset [16], [17]. This layer was selected because SAeUron provides a trained SAE for it and because prior work links it to object-level representations [7].

Consistent with the foundational literature on feature visualization, evaluation is qualitative because there is still no accepted metric for human interpretability [3], [4], [10], [12], [15]. To keep manual inspection feasible, calibration is performed on the first 30 channels of the raw layer and the first 30 SAE features. Each feature is generated with 5 fixed seeds for the prior and compared through three views: optimized visualizations, dataset examples assigned during time-step analysis (5 images per prompt across the UnlearnCanvas training set), and steering results.

**Pipeline stages** We run LVO as four sequential stages, each driven by its own configuration file: (i) **time-step activity analysis** captures top-$k$ activations across the dataset and produces an activity profile for every channel and SAE feature; (ii) **prior generation** steers the target feature on its active time-steps to obtain the prior image $P$; (iii) **visualization** initializes $z_{\text{opt}}$ from $\mathcal{E}(P)$ and runs Adam for 100 steps at learning rate 0.05 on a single A100 GPU (about a minute per feature); (iv) **evaluation** assembles the visualization, dataset examples, and steering results into the side-by-side views used in Section 5. Activation capture in stage (i) is the dominant cost and requires a multi-GPU sweep; SAE training is reused from prior work [7] and was not performed here. The full study, including all hyperparameter sweeps, took approximately 12 hours on 8 A100 GPUs on the PLGrid Cyfronet AGH cluster.

**Code and data** Source code, configuration files for each stage above, the fixed seeds used for the prior, and instructions for reproducing every stage of the pipeline are released at `github.com/aszokalski/diffusion-deep-dream-research`. The Stable Diffusion 1.5 weights fine-tuned on UnlearnCanvas [16], [17] and the SAEs from SAeUron [7] are obtained directly from the original releases.

## 4 Calibration

We tuned LVO's hyperparameters sequentially, in the spirit of coordinate descent [18]: a minimal baseline was fixed and one parameter group was tuned at a time, with the baseline updated only when a visually superior setting was found. The procedure follows the two stages of the pipeline described in Section 3: first the **prior generation** parameters (steering strength, set of time-steps to steer on), then the **visualization** parameters (the four classical regularizers and the two latent-domain penalties). Each parameter was swept over four levels per direction (e.g. weights 0, 0.5, 1, 5); full image grids for every sweep are provided in Section G. Schedule-matched noise was inconclusive across features and is therefore kept as an explicit LVO parameter rather than a fixed setting: every subsequent comparison is run with both `on` and `off` and the better of the two is reported per feature.

Five findings dominated calibration and motivate the analysis returned to in Section 6. (1) **Prior initialization was essential**: without it, both LVO variants collapsed into high-frequency noise on every sweep channel. (2) **Steering on active time-steps** was slightly but consistently better than steering on all time-steps, and the optimal scale differed sharply: 50 for raw-layer channels and 500 for SAE features, reflecting the sparser activation regime of the SAE. (3) **Spectral filtering** improved or tied nearly every comparison; **schedule-matched noise** was inconclusive (above). (4) **The two LVO variants prefer different regularization**: the raw-layer baseline benefited from mild total variation and gradient smoothing, while the SAE variant was hypersensitive to smoothing and benefited from a moment penalty instead. A light range penalty and mild transformation robustness improved both. (5) **Regularization can do more than suppress artifacts**: stronger transformation robustness and the moment penalty turned the muddy texture of SAE feature 14 into a clean cable-like structure consistent with its dataset examples (Section F), suggesting these terms also help the optimizer escape uninterpretable local maxima.

Table 1: Final hyperparameters selected for the two LVO variants after calibration. They share the same optimization core but require different regularization once SAE disentanglement is introduced.

| Parameter | Raw-layer | SAE |
|---|---|---|
| steering strength scale | 50 | 500 |
| steering time-steps | active_timesteps | active_timesteps |
| schedule noise | feature-dependent | feature-dependent |
| spectral filtering | on | on |
| learning rate | 0.05 | 0.05 |
| optimization steps | 100 | 100 |
| jitter / rotation / scale | 1 / 5° / 1.1 | 1 / 5° / 1.1 |
| total variation penalty | 0.5 | 0 |
| range penalty | 0.5 | 0.5 |
| moment penalty | 0 | 0.5 |
| gradient smoothing | 0.5 | 0 |

## 5 Results

We apply the calibrated LVO pipeline to features of the `up_blocks.1.attentions.1` layer of Stable Diffusion 1.5 (Style50). We present two raw-layer features as the polysemantic baseline and two SAE features as the monosemantic case studies. Extended activity and max-activation plots for all four are provided in the appendix.

### 5.1 Baseline: polysemantic visualizations

Optimizing raw layer activations produced results that were not entirely random, but most visualizations lacked clear interpretability and did not depict recognizable objects matching their dataset examples. The dataset examples themselves exhibited significant variety, consistent with these raw channels being polysemantic. We illustrate this with two representative features. Feature 25 mixes animal fragments - eyes, noses, paws, feathers, fur textures - across its activity peaks ($t = 61$, $t = 301$, $t = 421$); its dataset examples are exclusively animals (dogs, cats, bears) but never a single species. Feature 1214 produces butterfly-like shapes over unrelated backgrounds at $t = 1$, while the remaining peaks at $t = 181$ and $t = 401$ degenerate into seed-dependent noise.

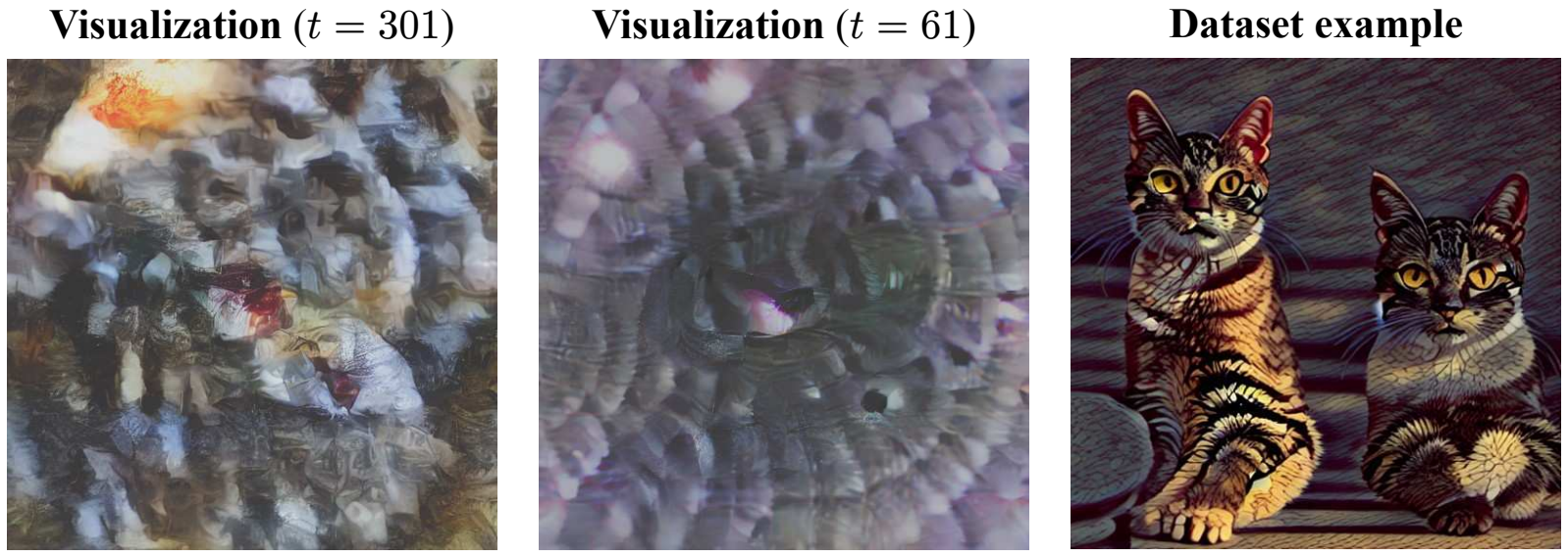

Figure 2: Raw feature 25. Visualizations mix fur and animal-part textures; dataset examples are exclusively animals (dogs, cats, bears) but no single species. Example prompt: *An Cats image in Pointillism style*.

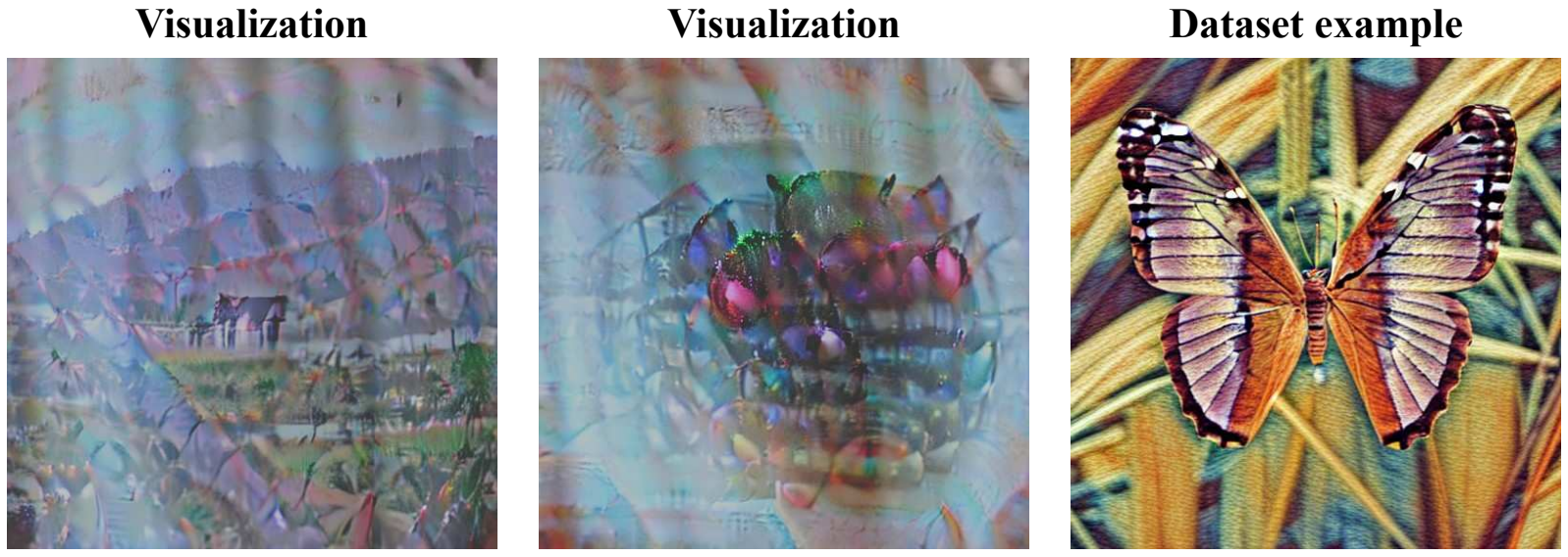

Figure 3: Raw feature 1214 at $t = 1$ across seeds. The earliest peak yields butterfly shapes; later peaks degenerate into noise that varies across seeds. Example prompt: *An Butterfly image in Pointillism style*.

For both features, steering produced uninformative images: a generic focal point in 1214′s case and no recognizable structure in 25′s. Peak activations (`14.3` for 25, `11.7` for 1214) are well above their normal operating ranges (`3-5` and `2.5-3.5` respectively), so the optimization is not failing at the activation level - it is finding adversarial maxima of polysemantic directions.

### 5.2 Monosemantic visualizations

Using the SAE to disentangle the features dramatically improved interpretability. We show that many SAE features produce clear and consistent visualizations across multiple activity peaks and random seeds. Their corresponding dataset examples depict more uniform concepts that match the visualizations well, suggesting that these features are genuinely more monosemantic than the raw-layer baseline.

#### 5.2.1 SAE feature 9984: diagonal composition

SAE feature 9984 produced visualizations showing delicate colorful strokes oriented from the top-left to the bottom-right of the image, consistently across all activity peaks.

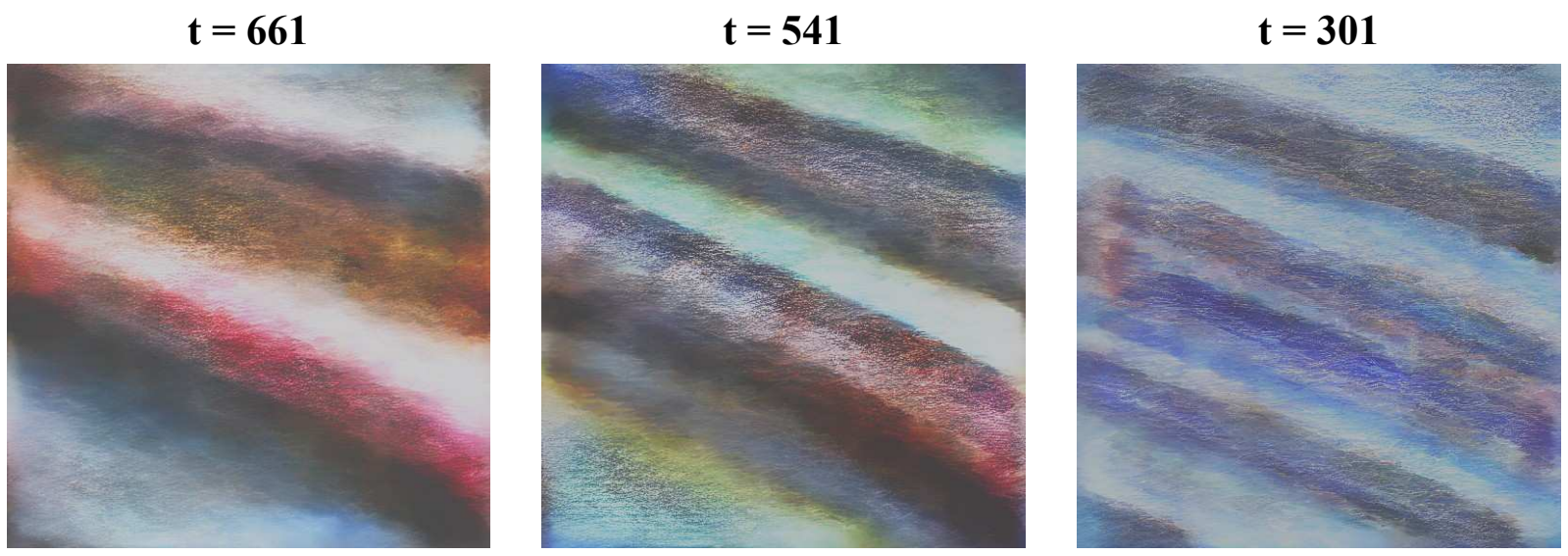

Figure 4: Selected visualizations of SAE feature 9984 at different time-steps and seeds.

The corresponding dataset examples clearly share a composition along the same diagonal direction. Notably, there is no common object or texture across these images; only the diagonal arrangement unifies them.

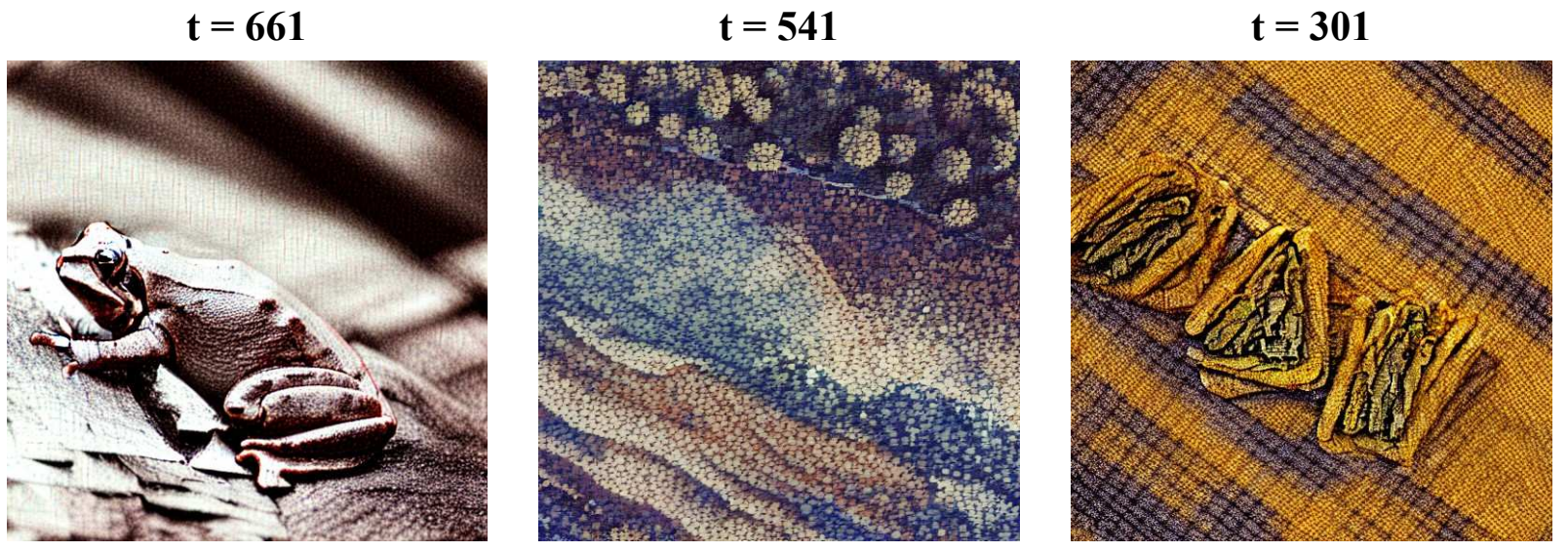

Figure 5: Selected dataset examples of SAE feature 9984 at different time-steps.

This feature appears monosemantic across all activity peaks. The strongest activation reached `10.6` at $t = 301$; at the other time-steps, the visualizations activated in the `7-8` range, which remains significant compared to the observed maximum activation of `3-4` during normal operation. Steering produced a more uniform gray-beige stroke pattern and was less informative than the optimized visualizations; the steered images are shown in Figure 8 in the appendix. These results provide strong evidence that this feature responds to diagonal compositional structure in the image. Analysing the dataset examples alone does not clearly reveal this interpretation, because the images vary substantially in content.

#### 5.2.2 SAE feature 10331: human figure

SAE feature 10331 produced visualizations clearly depicting a human figure, consistently across two of its three activity peaks and across variable seeds.

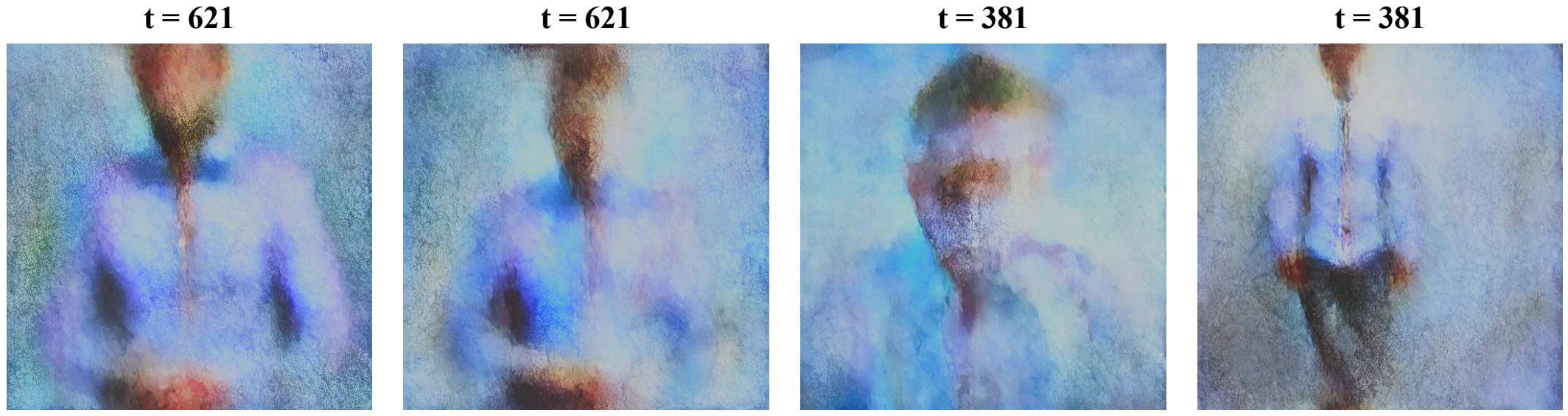

Figure 6: Selected visualizations of SAE feature 10331 at different time-steps and seeds.

The corresponding dataset examples also contain human figures rendered in various artistic styles.

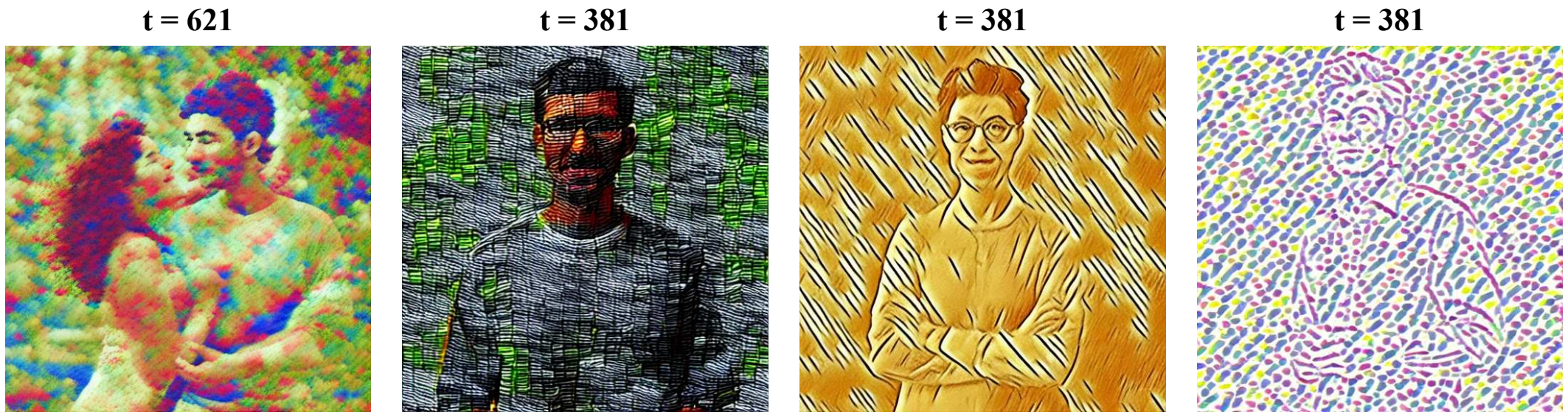

Figure 7: Selected dataset examples of SAE feature 10331 at different time-steps.

This feature appears monosemantic across two of its three activity peaks. The remaining peak at $t = 121$ produced plain images resembling the steering prior. The strongest activation reached `6.7` at $t = 621$; at $t = 381$ the activation was `3.4`, and at $t = 121$ it was `1.8`. Interestingly, the strongest activation was observed at the least active peak according to the frequency-in-top-k metric. This is an important reminder that time-step activity and maximum attainable activation are related but distinct signals. As with feature 9984 (Figure 8), steering mostly produced uniform gray-blue images and did not provide a meaningful explanation of the feature's function.

#### 5.2.3 Other monosemantic features

Beyond the two forementioned case studies, LVO recovered further monosemantic features at the same layer, including object parts (cables, roses), texture (waterfall foam), and composition (an inward perspective). We tabulate four representative examples—visualizations, dataset examples, and source prompts—in Table 2 in the appendix.

## 6 Discussion

**RQ1** We show that LVO is viable in the LDM latent space when its four components are combined: time-step activity analysis, schedule-matched noise, prior initialization, and appropriate regulariza-tion. The time-step analysis is essential: different peaks produced different visualization quality, and (per feature 10331) the highest activation did not always occur on the most active peak.

**RQ2** We demonstrate that the baseline produces polysemantic visualizations that cannot be mapped to single concepts, whereas LVO with SAE disentanglement yields features consistent across activity peaks and seeds, correlating with much more uniform dataset examples. This is also a statement about the quality of the SAE: it disentangles the layer into meaningful features.

**RQ3** We show that pixel-space regularizers transfer but do not behave identically; in latent space they also guide the optimization toward coherent concepts. The raw-layer variant of LVO benefited from total variation and gradient smoothing; the SAE variant preferred neither, and instead benefited from a moment penalty.

**RQ4** LVO complements dataset examples and steering. Dataset examples are limited by content (the diagonal composition of 9984 is not obvious from examples); steering reveals downstream effects, not the cause of activation, and was uninformative here.

## 7 Limitations

Evaluation is qualitative because no accepted metric for human interpretability exists, consistent with the foundational feature- visualization literature [3], [4], [10], [12], [15]; a quantitative study is left for future work. The empirical scope is one model (SD 1.5 fine-tuned on Style50), one layer (`up1.1`), and the first `30` raw and SAE features. Individual features may still be multifaceted [12], and SAE-based results are upper-bounded by the disentanglement quality of the SAE itself.

## 8 Broader impacts

Feature-level interpretability of generative image models has a primarily defensive profile: it supports safety auditing, targeted concept **unlearning** [7], [16], content moderation, and transparency for regulators. Monosemantic feature visualization is also a prerequisite for extending **circuit analysis** [5] to latent diffusion: identifying which features in earlier layers excite a downstream feature, and tracing how concepts compose through the U-Net—which we see as the most productive next step for mechanistic study of LDMs. The dual-use dimension is that an actor who can identify which features a safety classifier or unlearning method relies on could in principle steer those features to bypass filters without changing the prompt. We assess this risk as limited here because the SAEs and Style50 model are already public [7], LVO shows what activates a feature rather than how to suppress one beyond existing steering work [7], [13], and we release only the analysis framework —no new models or circumvention tools.

## 9 Conclusion

We have proposed **latent visualization by optimization** (LVO) and demonstrated that it brings feature visualization by optimization to latent diffusion models. The most striking case in our study is SAE feature 9984: its dataset examples vary so widely in content that no shared concept is legible by inspection, yet LVO consistently renders diagonal compositions across seeds and time-steps. LVO surfaces structure that dataset examples and steering—the two existing interpretability tools for diffusion models—do not. It thus fills a complementary niche that diffusion-model interpretability has so far been missing.

## Acknowledgments

We gratefully acknowledge Poland's high-performance Infrastructure PLGrid ACK Cyfronet AGH for providing computer facilities and support within computational grant no. PLG/2025/018892.

## References

[1] S. Russell, *Human Compatible: Artificial Intelligence and the Problem of Control*. Viking, 2019.

[2] C. Olah, “Mechanistic Interpretability, Variables, and the Importance of Interpretable Bases,” *Transformer Circuits Thread*, 2022.

[3] A. Mordvintsev, C. Olah, and M. Tyka, “Going deeper into neural networks.” [Online]. Available: https://research.google/blog/inceptionism-going-deeper-into-neural-networks/

[4] C. Olah, A. Mordvintsev, and L. Schubert, “Feature Visualization,” *Distill*, 2017, doi: 10.23915/distill.00007.

[5] C. Olah, N. Cammarata, L. Schubert, G. Goh, M. Petrov, and S. Carter, “Zoom In: An Introduction to Circuits,” *Distill*, 2020, doi: 10.23915/distill.00024.001.

[6] T. Bricken *et al.*, “Towards Monosemanticity: Decomposing Language Models With Dictionary Learning,” *Transformer Circuits Thread*, 2023.

[7] B. Cywiński and K. Deja, “SAeUron: Interpretable Concept Unlearning in Diffusion Models with Sparse Autoencoders.” [Online]. Available: https://arxiv.org/abs/2501.18052

[8] J. Ho, A. Jain, and P. Abbeel, “Denoising Diffusion Probabilistic Models.” [Online]. Available: https://arxiv.org/abs/2006.11239

[9] R. Rombach, A. Blattmann, D. Lorenz, P. Esser, and B. Ommer, “High-Resolution Image Synthesis with Latent Diffusion Models.” [Online]. Available: https://arxiv.org/abs/2112.10752

[10] A. Odena, V. Dumoulin, and C. Olah, “Deconvolution and Checkerboard Artifacts,” *Distill*, 2016, doi: 10.23915/distill.00003.

[11] A. M. Nguyen, A. Dosovitskiy, J. Yosinski, T. Brox, and J. Clune, “Synthesizing the preferred inputs for neurons in neural networks via deep generator networks,” *CoRR*, 2016, [Online]. Available: http://arxiv.org/abs/1605.09304

[12] A. Nguyen, J. Yosinski, and J. Clune, “Multifaceted Feature Visualization: Uncovering the Different Types of Features Learned By Each Neuron in Deep Neural Networks.” [Online]. Available: https://arxiv.org/abs/1602.03616

[13] A. Templeton *et al.*, “Scaling Monosemanticity: Extracting Interpretable Features from Claude 3 Sonnet,” *Transformer Circuits Thread*, 2024, [Online]. Available: https://transformer-circuits.pub/2024/scaling-monosemanticity/index.html

[14] A. Mahendran and A. Vedaldi, “Understanding Deep Image Representations by Inverting Them.” [Online]. Available: https://arxiv.org/abs/1412.0035

[15] A. M. Øygard, “Visualizing GoogLeNet Classes.” 2015.

[16] OPTML-Group, “Unlearn Canvas: Style50 fine-tuned model.” GitHub, 2018.

[17] Y. Zhang *et al.*, “UnlearnCanvas: Stylized Image Dataset for Enhanced Machine Unlearning Evaluation in Diffusion Models.” [Online]. Available: https://arxiv.org/abs/2402.11846

[18] J. Bergstra and Y. Bengio, “Random Search for Hyper-Parameter Optimization,” *Journal of Machine Learning Research*, vol. 13, no. 10, pp. 281–305, 2012, [Online]. Available: http://jmlr.org/papers/v13/bergstra12a.html

[19] C. Olah, N. Cammarata, L. Schubert, G. Goh, M. Petrov, and S. Carter, “An Overview of Early Vision in InceptionV1,” *Distill*, 2020, doi: 10.23915/distill.00024.002.

## A Steering produces uninformative images

The main text claims that steering reveals downstream effects rather than the cause of activation, and was uninformative for the features studied here. Figure 8 illustrates this for SAE feature 9984 (the “diagonal composition” feature of Section 5.2.1). The optimized visualization shows clear oriented strokes; the steered images do not display any consistent diagonal structure and look like generic unconditional generations.

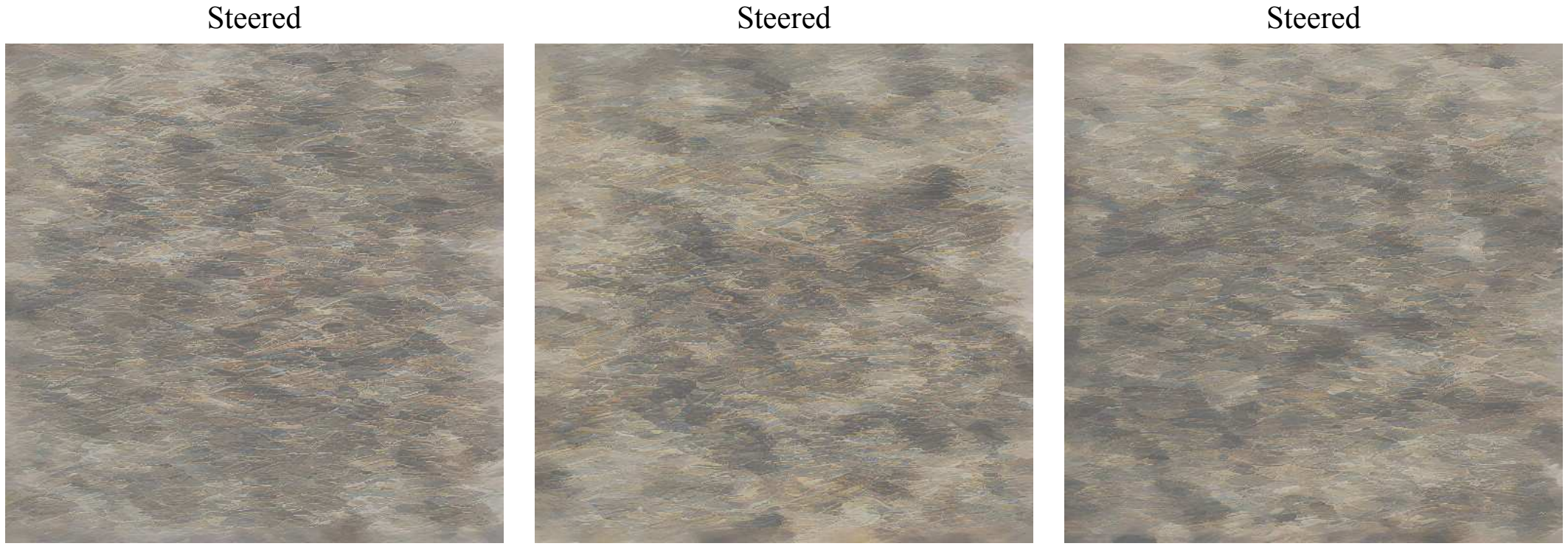

Figure 8: Steering results for SAE feature 9984. The steered images do not exhibit the diagonal composition that the optimization-based visualization (Figure 4) clearly reveals.

## B Other monosemantic features

Table 2 extends the qualitative results of Section 5.2.3 to four additional SAE features. They show that LVO recovers object-part, texture, and composition features, not only the two flagship cases discussed in the main text.

Table 2: Other monosemantic SAE features identified using LVO.

| Feature | Description | Visualization | Dataset Example | Example prompt[1] |
|---|---|---|---|---|
| 14 | cables | | | An Butterfly image in Pointillism style |
| 10204 | roses | | | An Flowers image in Winter style |
| 10272 | inward perspective | | | An Towers image in Surrealism style |
| 10538 | foam | | | An Waterfalls image in French style |

## C Extended activity diagnostics

[1]These prompts are taken from the UnlearnCanvas dataset and are not grammatically corrected.

The plots in Figure 9 and Figure 10 provide the activity diagnostics for the four main case studies discussed in the paper. They support two claims from the main text: first, activity peaks are a useful way to reduce the temporal complexity of diffusion models; second, activity frequency and maximum attainable activation are not identical signals.

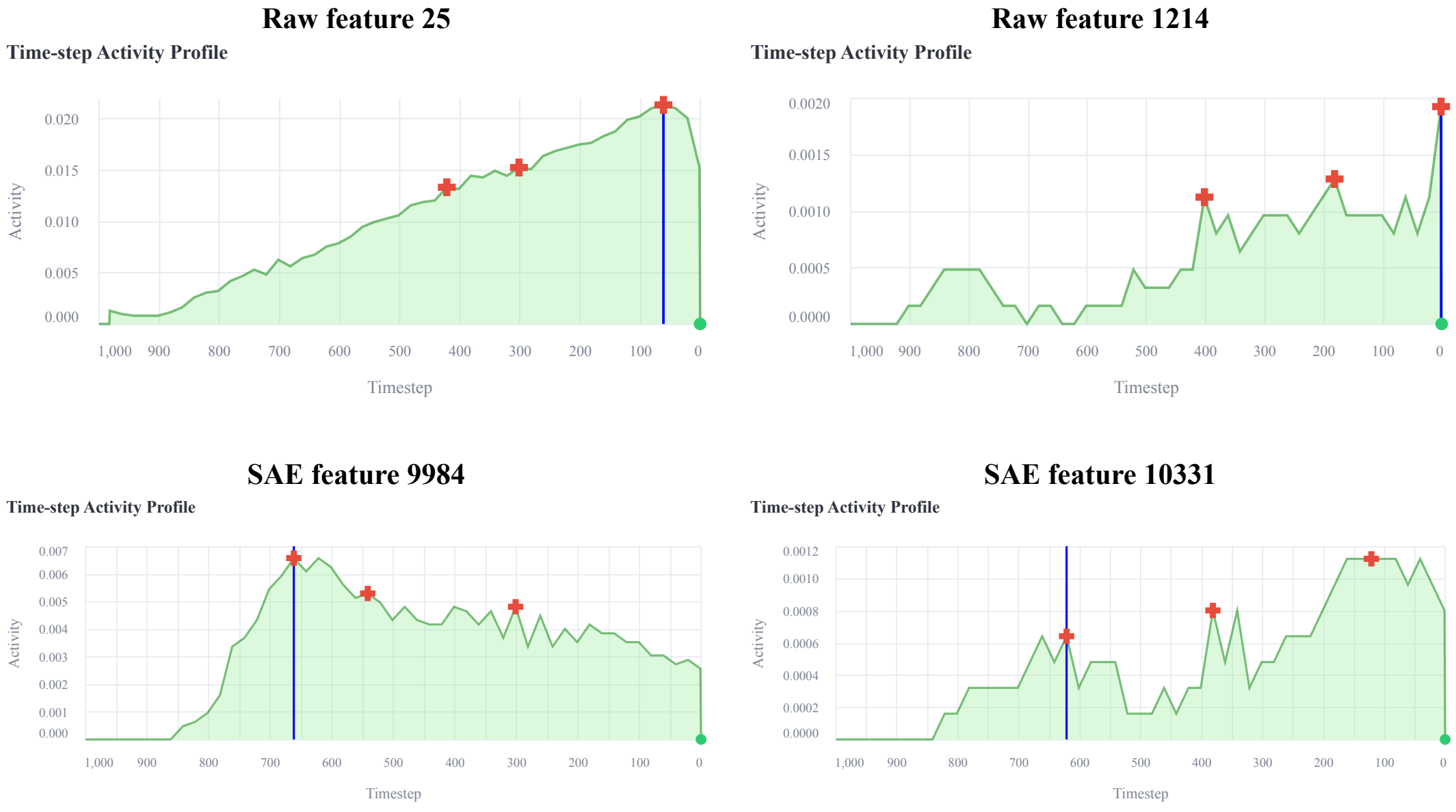

Figure 9: Time-step activity profiles for the four main case studies. Activity is the frequency with which a feature appears among the top-k activations at a given time-step across the dataset.

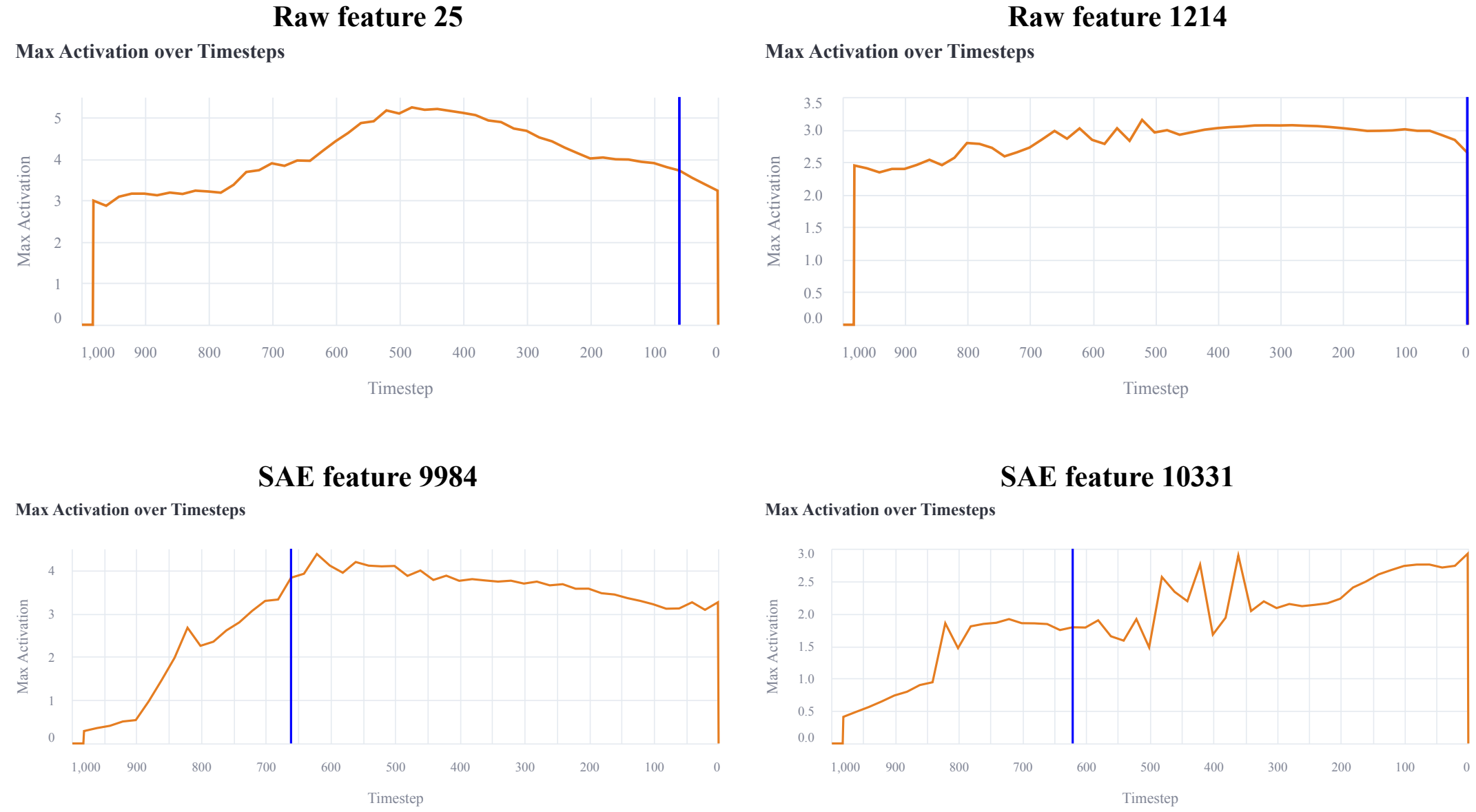

Figure 10: Maximum activation over time-steps for the four main case studies. For feature 10331, the strongest visualization occurs on a less active peak, illustrating that activity and maximum activation are distinct diagnostics.

## D Early-layer sanity checks

Before introducing the full regularization stack, we tested a naive version of LVO on early convolutional layers of Stable Diffusion 1.5. Even without an SAE, the visualizations were clear and interpretable because optimizing early layers does not require backpropagating through many convolutional blocks, which are a major source of adversarial high-frequency noise [10]. The results resemble the early-layer detectors reported for `InceptionV1`, including geometric and color-sensitive patterns [19].

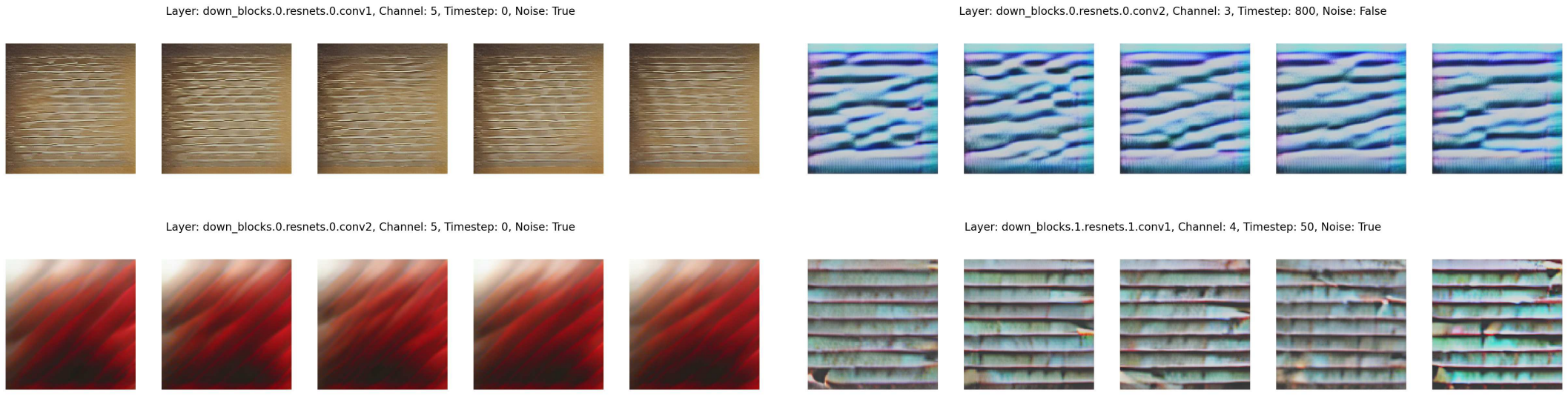

Figure 11: Early-layer visualizations from Stable Diffusion 1.5 generated with five seeds. Their similarity to classical CNN feature visualizations supports the view that some low-level features are architecture-independent.

## E Schedule-matched noise injection examples

Schedule-matched noise injection improved some visualizations and degraded others. This behavior appeared repeatedly during calibration and motivated treating noise injection as a feature-dependent switch rather than a universal default.

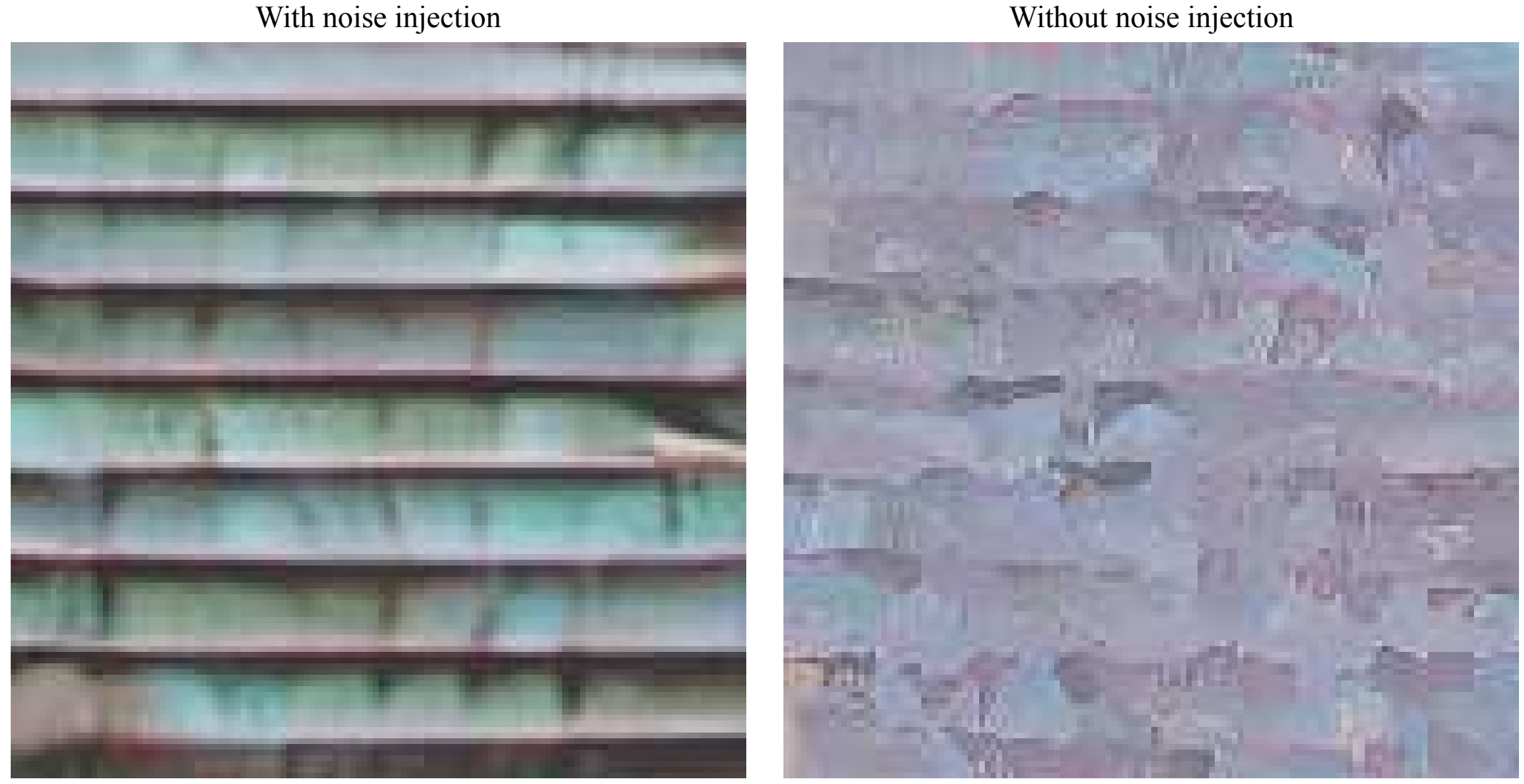

Figure 12: Example of improvement with noise injection: layer `down_blocks.1.resnets.1.conv1`, channel `4`, time-step `50`.

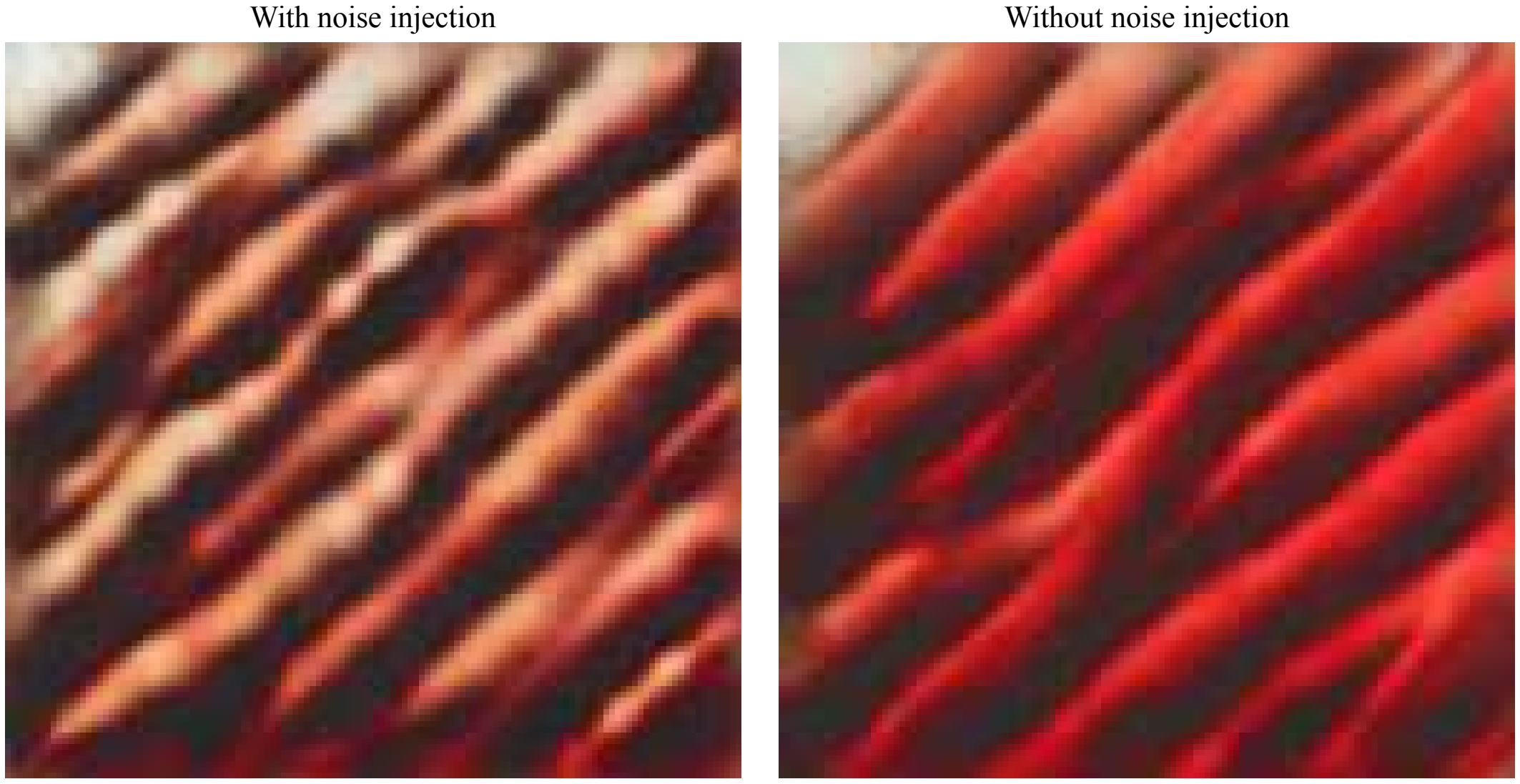

Figure 13: Example of degradation with noise injection: layer `down_blocks.1.resnets.1.conv1`, channel `5`, time-step `200`.

## F Cable-like structure discovered during calibration

One of the most interesting findings from calibration was the behavior of SAE feature 14. Under weak regularization it produced a muddy crowd-like texture, but stronger transformation robustness revealed a cable-like structure that was also present in dataset examples. This suggests that regularization can do more than suppress artifacts: it can help the optimizer escape an uninterpretable local maximum and reveal a more meaningful facet of the feature.

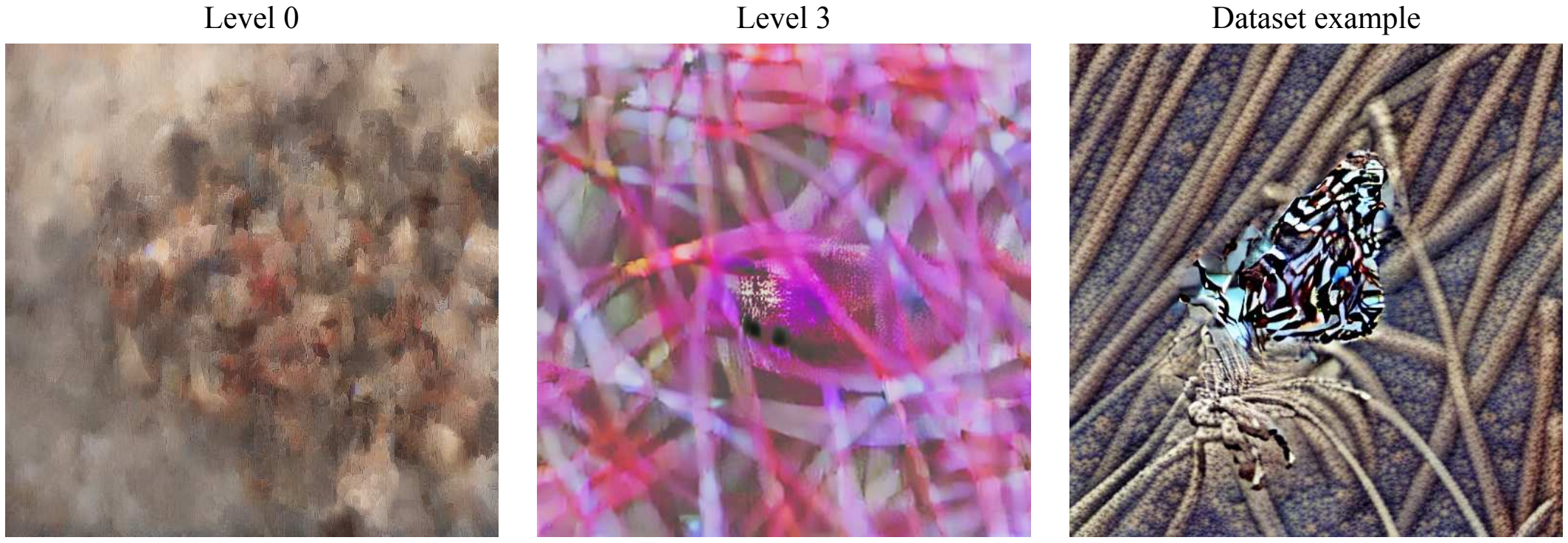

Figure 14: Transformation robustness changes SAE feature 14 from an uninterpretable texture into a cable-like structure consistent with the training data.

## G Calibration sweeps

This section collects the qualitative image grids that drove each coordinate-descent step in Section 4. Subsections follow the order of the calibration in the underlying study: prior generation parameters first (steering strength, set of time-steps), then visualization parameters (prior regularization, schedule-matched noise, spectral filtering, learning rate and optimization steps, transformation robustness, and the four penalties). Within each table, rows index the swept value (or noise condition for sweeps where schedule noise is also varied) and columns index the channel or SAE feature.

### G.1 Steering strength scale

We swept `steering_strength_scale` $\in \{1, 5, 10, 50, 100, 1000\}$ for both methods, with an additional `500` for the SAE method when the trends from the smaller values suggested a shift in the effective range.

Table 3: Steering strength sweep, raw-layer method.

| Value | ch2 | ch3 | ch5 | ch6 | ch8 | ch10 |
|---|---|---|---|---|---|---|
| 1 | | | | | | |
| 5 | | | | | | |
| 10 | | | | | | |
| 50 | | | | | | |
| 100 | | | | | | |
| 1000 | | | | | | |
| **Best** | 50 | 50 | 10/50 | 100 | 50 | 50 |

| Value | ch19 | ch21 | ch22 | ch25 | ch26 | ch30 |
|---|---|---|---|---|---|---|
| 1 | | | | | | |
| 5 | | | | | | |

| Value | ch19 | ch21 | ch22 | ch25 | ch26 | ch30 |
|---|---|---|---|---|---|---|
| 10 | | | | | | |
| 50 | | | | | | |
| 100 | | | | | | |
| 1000 | | | | | | |
| **Best** | 100 | 50 | 50 | 100 | 100 | 100 |

For the raw-layer method, `steering_strength_scale = 50` was optimal. Smaller values failed to produce meaningful variations: the only seemingly meaningful pattern (a house-like structure on `ch5` at strength `10`) was a seed-dependent artifact, since the same pattern appeared on unrelated channels under the same seed. A value of `1000` produced either pure noise or static color fields, neither suitable as a prior. Values of `50` and `100` produced coherent structures resembling leaves (`ch5`, `ch6`), textures (`ch22`, `ch25`), and crowds (`ch2`); we selected `50` for the balance between structural coherence and ambiguity - high enough to generate a recognisable object prior, low enough not to constrain the optimization to merely reproduce that prior.

Table 4: Steering strength sweep, SAE method.

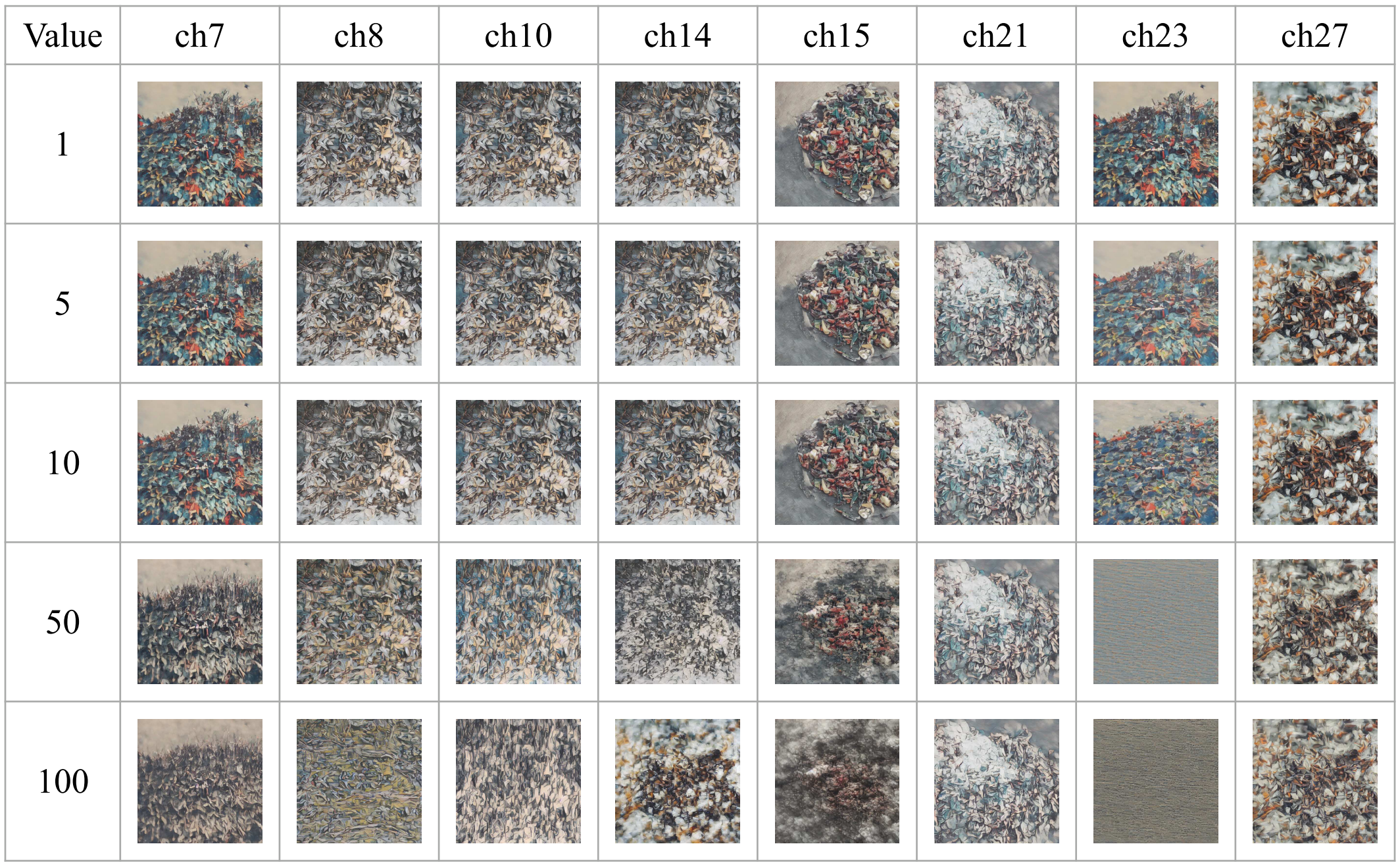

| Value | ch7 | ch8 | ch10 | ch14 | ch15 | ch21 | ch23 | ch27 |
|---|---|---|---|---|---|---|---|---|
| 1 | | | | | | | | |
| 5 | | | | | | | | |
| 10 | | | | | | | | |
| 50 | | | | | | | | |
| 100 | | | | | | | | |

| Value | ch7 | ch8 | ch10 | ch14 | ch15 | ch21 | ch23 | ch27 |
|---|---|---|---|---|---|---|---|---|
| 500 | | | | | | | | |
| 1000 | | | | | | | | |
| **Best** | 500/1000 | 500 | 100 | 500 | 100 | 500 | 100 | 1000 |

For the SAE method, `steering_strength_scale = 500` was optimal. Steering required substantially larger magnitudes to produce meaningful changes, consistent with the inherent sparsity of SAE activations. Most channels showed no impact until values reached `500-1000`; `1000` produced high-contrast results visually analogous to `100` in the raw-layer regime (e.g., `ch27` rendered a clear flowers image). We chose `500` for the same balance reason as in the raw-layer case. After this step we restricted subsequent sweeps to channels with the most promising priors: `{19, 21, 22, 25, 26, 30}` for the raw-layer method and `{10, 14, 15, 21, 23, 27}` for the SAE method.

### G.2 Time-steps to steer on

We compared steering on `active_timesteps` (only the time-steps with non-zero activity in the feature's profile) versus `all_timesteps`.

Table 5: Active vs. all time-step steering, raw-layer method.

| Timesteps | ch19 | ch21 | ch22 | ch25 | ch26 | ch30 |
|---|---|---|---|---|---|---|
| Active | 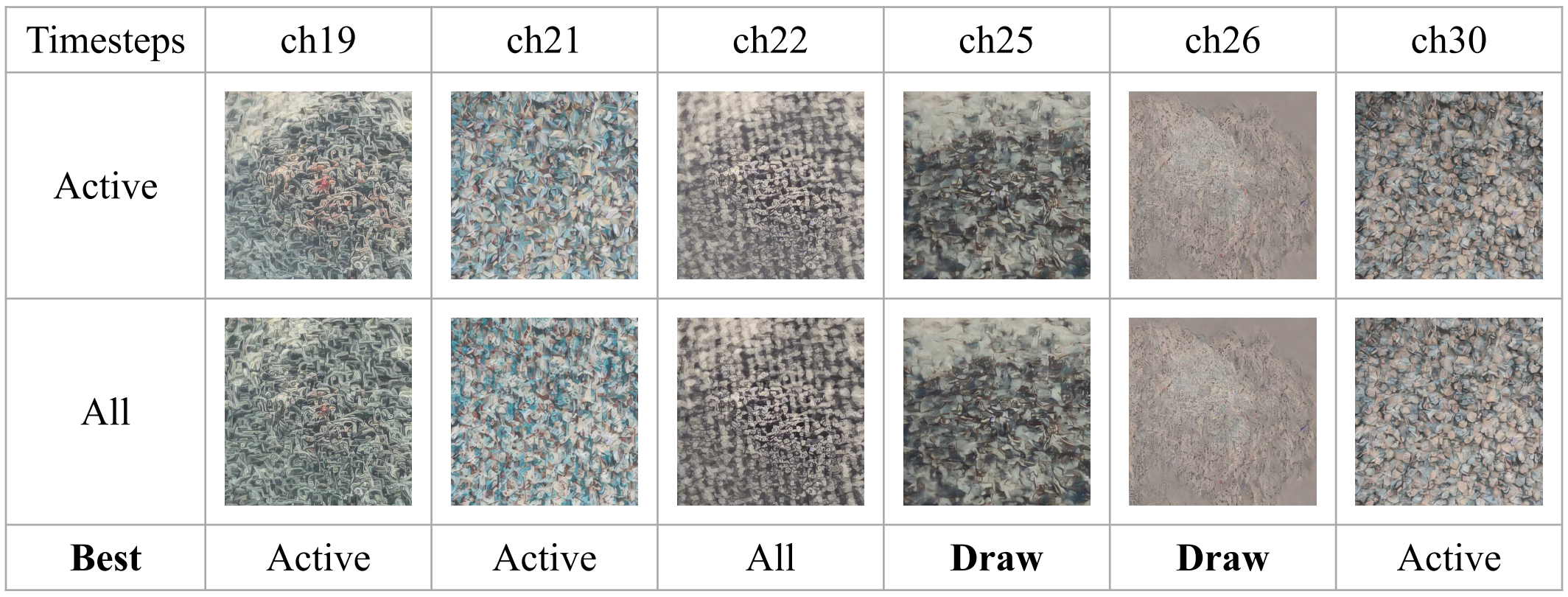 | | | | | |
| All | | | | | | |
| **Best** | Active | Active | All | **Draw** | **Draw** | Active |

Table 6: Active vs. all time-step steering, SAE method.

| Timesteps | ch10 | ch14 | ch15 | ch21 | ch23 | ch27 |
|---|---|---|---|---|---|---|
| Active | 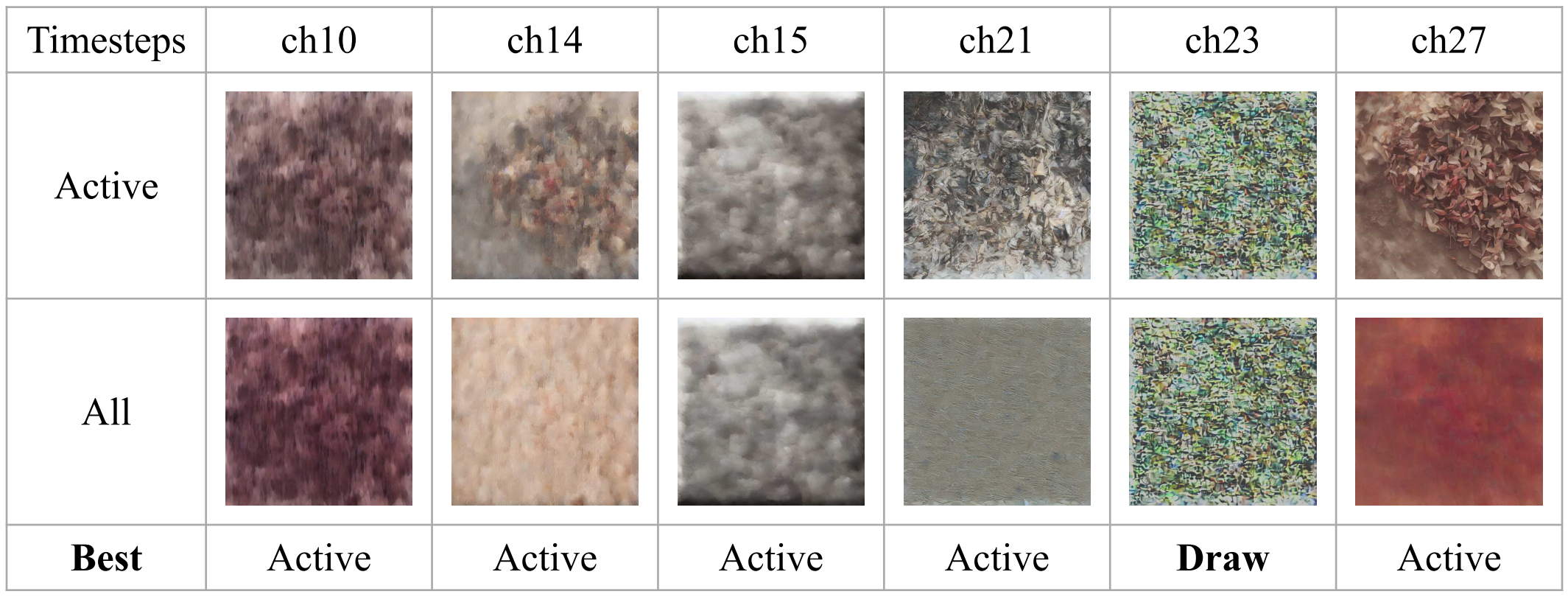 | | | | | |
| All | | | | | | |
| **Best** | Active | Active | Active | Active | **Draw** | Active |

The differences were subtle but consistent: active-time-step steering was less washed out (`ch22`), produced more recognisable objects (a bird on `ch19`), and avoided the static-colour collapse seen on `ch27`. For the draw cases on raw-layer `ch25` and `ch26` the active set essentially matched the full set, since their activity profiles were non-zero across the entire range. We set `timesteps = active_timesteps` for both methods.

### G.3 Prior regularization

This sweep tested whether feature-steered prior initialization improves results over random-noise initialization.

Table 7: Prior vs. random-noise initialization, raw-layer method.

| Init | ch19 | ch21 | ch22 | ch25 | ch26 | ch30 |
|---|---|---|---|---|---|---|
| No Prior | | | | | | |
| Prior | | | | | | |
| **Best** | Prior | Prior | Prior | Prior | Prior | Prior |

For the raw-layer method, prior initialization consistently outperformed random-noise initialization across all tested channels. Without a prior, the optimization produced dense pixel-level fragmentation with no semantic structure - the adversarial-artifact failure mode that regularization is meant to mitigate. With a prior, `ch19` showed trees and hills, `ch21` smooth brushstroke textures with clear colour gradients, `ch22` a recognisable wood texture with a cavity, `ch25` feathers and bird-like structures, and `ch30` flowers.

Table 8: Prior vs. random-noise initialization, SAE method.

| Init | ch10 | ch14 | ch15 | ch21 | ch23 | ch27 |
|---|---|---|---|---|---|---|
| No Prior | | | | | | |
| Prior | | | | | | |
| **Best** | Prior | Prior | Prior | Prior | Prior | Prior |

The SAE method showed the same pattern: every random-noise channel was indistinguishable from noise, while prior initialization produced coherent results (`ch10` shifted to soft reddish-magenta tones, `ch15` to coherent diagonal green strokes, `ch21` to leaf-like shapes, `ch27` to faded floral compositions). Prior regularization is therefore enabled by default for both methods.

### G.4 Schedule-matched noise injection

For each channel we optimized at the first activity peak and compared results with and without schedule-matched noise injection.

Table 9: Schedule-matched noise on/off, raw-layer method.

| Setting | ch19 | ch21 | ch22 | ch25 | ch26 | ch30 |
|---|---|---|---|---|---|---|
| No Noise | | | | | | |
| Noise | | | | | | |
| Time-step | 161 | 541 | 321 | 301 | 1 | 561 |
| **Best** | Noise | No Noise | **Draw** | No Noise | **Draw** | Noise |

For the raw-layer method, noise injection improved `ch19` and `ch30` (clearer hills/trees and clearer foliage respectively), while `ch21` and `ch25` were better without noise (smoother textures and a cleaner fur-and-nose pattern). `ch22` and `ch26` showed no clear winner.

Table 10: Schedule-matched noise on/off, SAE method.

| Setting | ch10 | ch14 | ch15 | ch21 | ch23 | ch27 |
|---|---|---|---|---|---|---|
| No Noise | | | | | | |
| Noise | | | | | | |
| Time-step | 641 | 1 | 161 | 1 | 41 | 561 |
| **Best** | **Draw** | **Draw** | Noise | **Draw** | **Draw** | No Noise |

The SAE method was similarly inconclusive: most channels (`ch10`, `ch14`, `ch21`, `ch23`) produced nearly identical results in both conditions; `ch15` improved slightly with noise (clearer green diagonal strokes) and `ch27` was better without noise (more coherent pink/red flower pattern). Since neither setting consistently outperformed the other - and there is no theoretical reason to expect raw-layer and SAE methods to behave differently with respect to noise injection - we treat `see_through_schedule_noise` as an explicit method parameter and evaluate both configurations in subsequent sweeps.

### G.5 Gradient spectral filtering

We compared optimization with and without spectral filtering, for both schedule-noise conditions.

Table 11: Spectral filtering on/off across both noise conditions, raw-layer method.

| **Noise** | **Setting** | **ch19** | **ch21** | **ch22** | **ch25** | **ch26** | **ch30** |
|---|---|---|---|---|---|---|---|
| No Noise | Off | | | | | | |

| Noise | Setting | ch19 | ch21 | ch22 | ch25 | ch26 | ch30 |
|---|---|---|---|---|---|---|---|
| | On | | | | | | |
| Noise | Off | | | | | | |
| | On | | | | | | |
| **Best** | | On | Off | On | On | Draw | On |

For the raw-layer method, spectral filtering produced better results in the majority of comparisons. `ch19` and `ch22` showed clearer hills/holes and cleaner colours respectively, and `ch30` showed cleaner leaves under the noise condition. The exception was `ch21`, where unfiltered runs preserved more colour detail and the filtered runs looked washed out.

Table 12: Spectral filtering on/off across both noise conditions, SAE method.

| Noise | Setting | ch10 | ch14 | ch15 | ch21 | ch23 | ch27 |
|---|---|---|---|---|---|---|---|
| No Noise | Off | | | | | | |
| | On | | | | | | |
| Noise | Off | | | | | | |
| | On | | | | | | |
| **Best** | | Draw | Draw | On | On | Draw | On |

For the SAE method, spectral filtering won or drew in all comparisons, with no losses: `ch15` produced smoother diagonal strokes (clearer green under noise), `ch27` produced cleaner flower patterns, and the remaining channels showed no significant difference. Spectral filtering is enabled by default for both methods.

### G.6 Learning rate and optimization steps

We swept learning rate $\in \{0.001, 0.01, 0.05, 0.1\}$ against optimization steps $\in \{100, 200, 1000, 2000\}$ on two representative channels per method (`ch19` and `ch25` for the raw-layer method, `ch15` and `ch27` for the SAE method).

Table 13: Learning rate $\times$ optimization-steps sweep for raw-layer ch19, with schedule-matched noise.

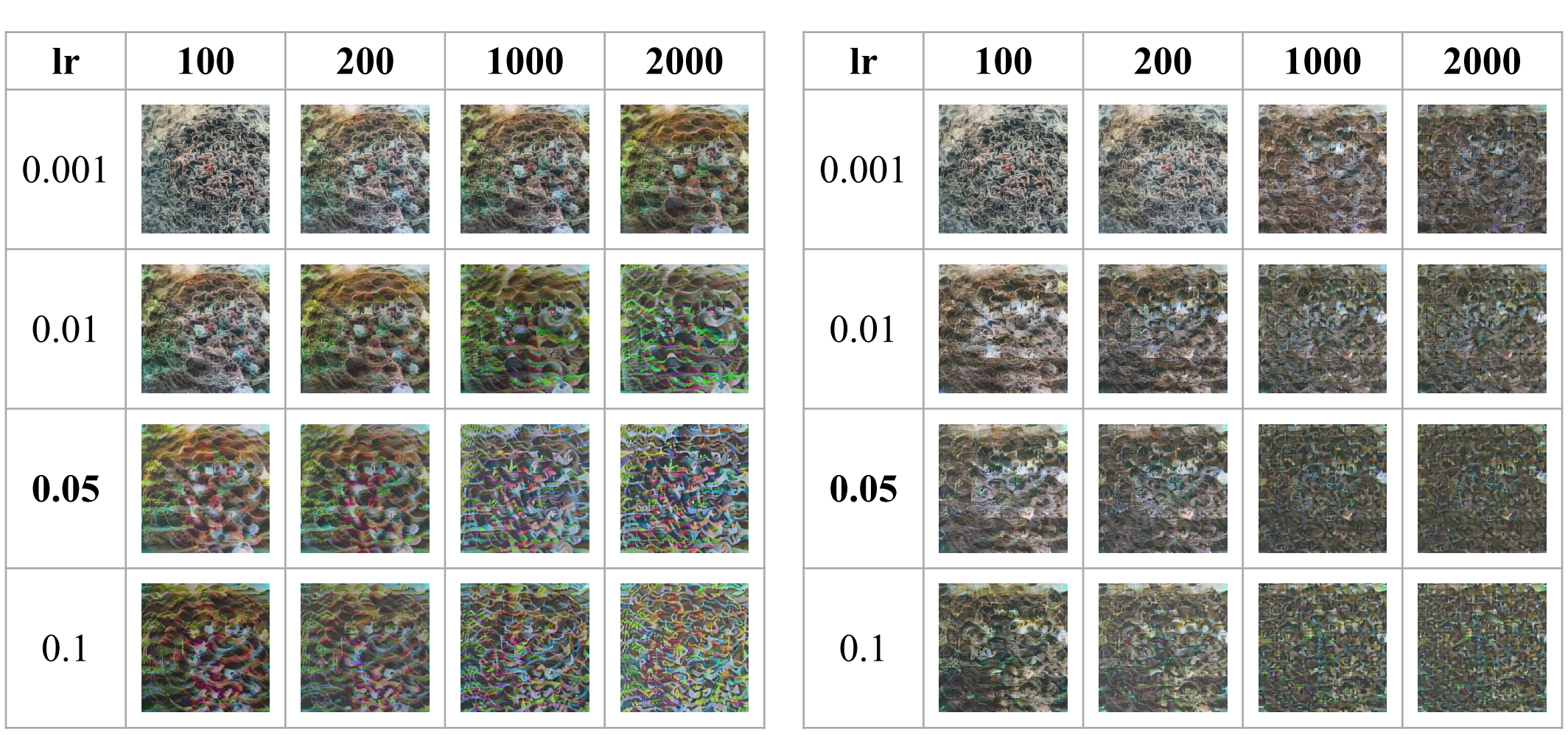

**Noise**

| lr | 100 | 200 | 1000 | 2000 |
|---|---|---|---|---|
| 0.001 | | | | |
| 0.01 | | | | |
| **0.05** | | | | |
| 0.1 | | | | |

**No Noise**

| lr | 100 | 200 | 1000 | 2000 |
|---|---|---|---|---|
| 0.001 | | | | |
| 0.01 | | | | |
| **0.05** | | | | |
| 0.1 | | | | |

Table 14: Learning rate × optimization-steps sweep for raw-layer ch25, with schedule-matched noise.

**Noise**

| lr | 100 | 200 | 1000 | 2000 |
|---|---|---|---|---|
| 0.001 | | | | |
| 0.01 | | | | |
| **0.05** | | | | |
| 0.1 | | | | |

**No Noise**

| lr | 100 | 200 | 1000 | 2000 |
|---|---|---|---|---|
| 0.001 | | | | |
| 0.01 | | | | |
| **0.05** | | | | |
| 0.1 | | | | |

The pattern was clear. Very small learning rates (`0.001`) did not converge in `100-200` steps. Higher rates (`0.05`, `0.1`) converged at `100` steps; running for `1000-2000` steps caused over-optimization with visible high-frequency artifacts. For `ch19`, `lr=0.05` or `0.1` with `100` steps produced clear hills and trees; `lr=0.01` with `1000` steps also produced coherent structure, showing that smaller learning rates work given more steps. For `ch25`, `lr=0.05` with `100` steps was best.

Table 15: Learning rate × optimization-steps sweep for SAE ch15, with schedule-matched noise.

**Noise**

| lr | 100 | 200 | 1000 | 2000 |
|---|---|---|---|---|
| 0.001 | | | | |
| 0.01 | | | | |

**No Noise**

| lr | 100 | 200 | 1000 | 2000 |
|---|---|---|---|---|
| 0.001 | | | | |
| 0.01 | | | | |

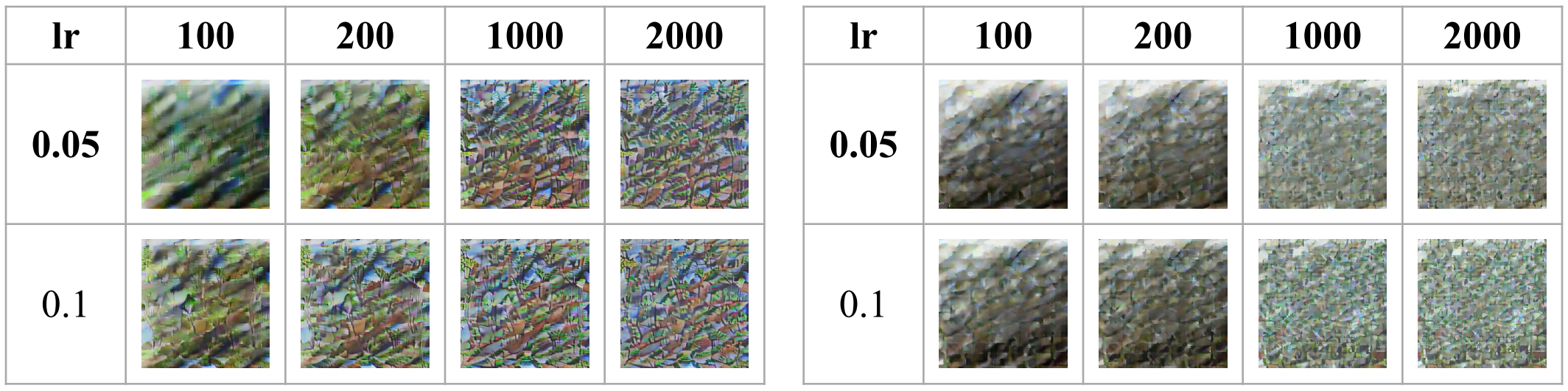

| lr | 100 | 200 | 1000 | 2000 |
|---|---|---|---|---|
| **0.05** | | | | |
| 0.1 | | | | |

| lr | 100 | 200 | 1000 | 2000 |
|---|---|---|---|---|
| **0.05** | | | | |
| 0.1 | | | | |

Table 16: Learning rate × optimization-steps sweep for SAE ch27, without schedule-matched noise.

**Noise** **No Noise**

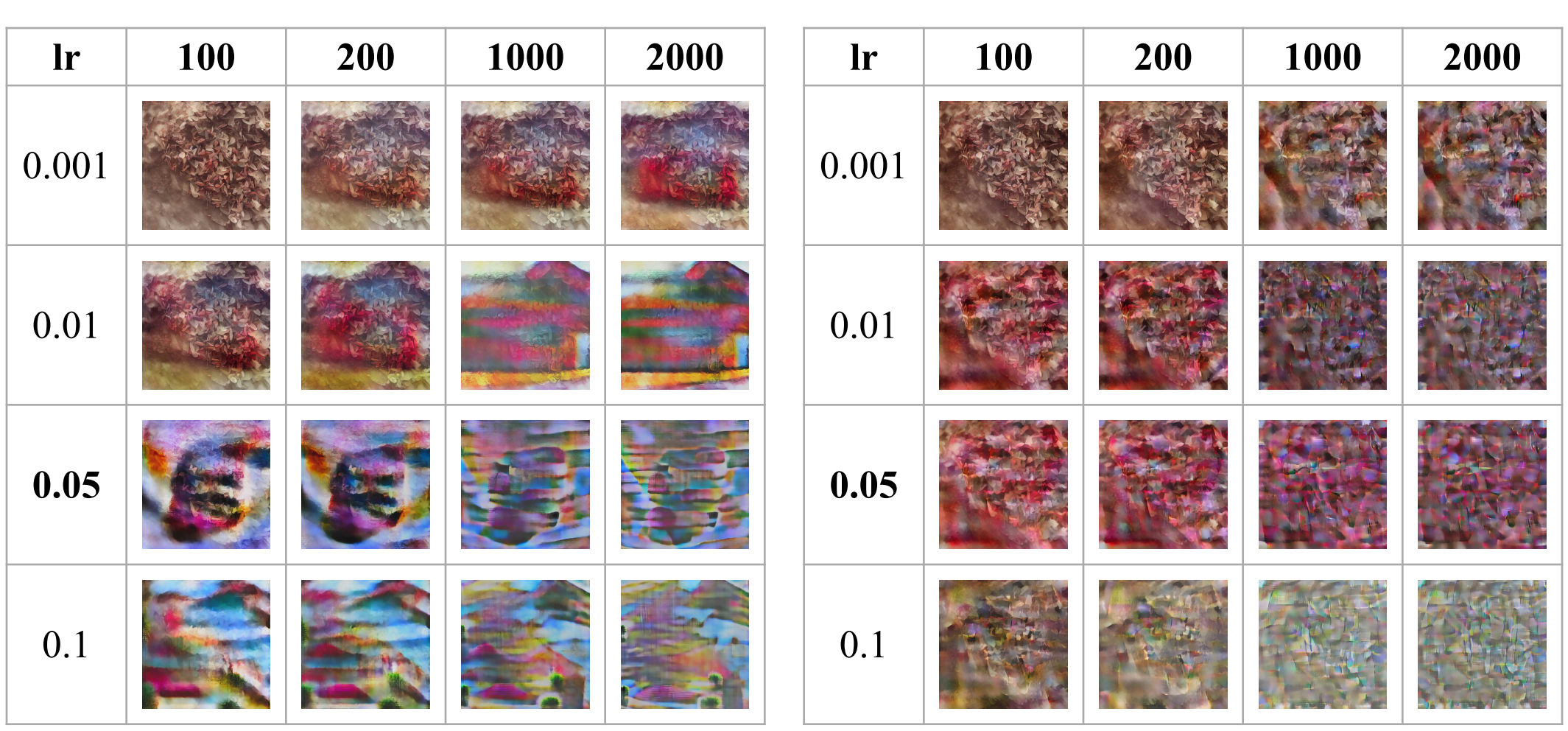

| lr | 100 | 200 | 1000 | 2000 |
|---|---|---|---|---|
| 0.001 | | | | |
| 0.01 | | | | |
| **0.05** | | | | |
| 0.1 | | | | |

| lr | 100 | 200 | 1000 | 2000 |
|---|---|---|---|---|
| 0.001 | | | | |
| 0.01 | | | | |
| **0.05** | | | | |
| 0.1 | | | | |

The SAE method showed similar patterns: `lr=0.05` with `100-200` steps produced clean diagonal stroke patterns on `ch15`, and `lr=0.05` with `100` steps produced clean flower patterns on `ch27`. We selected `lr=0.05` with `100` steps as the default for both methods - fast convergence without over-optimization artifacts.

### G.7 Transformation robustness

We compared four robustness levels: 0 (disabled), 1 (jitter `1`, rotation `5°`, scale `1.1`), 2 (`8`, `15°`, `1.2`), and 3 (`16`, `45°`, `1.8`).

Table 17: Transformation-robustness sweep, raw-layer method. Levels increase jitter, rotation, and scale jointly.

| Noise | Level | ch19 | ch21 | ch22 | ch25 | ch26 | ch30 |
|---|---|---|---|---|---|---|---|
| No Noise | 0 | | | | | | |
| | 1 | | | | | | |
| | 2 | | | | | | |

| Noise | Level | ch19 | ch21 | ch22 | ch25 | ch26 | ch30 |
|---|---|---|---|---|---|---|---|
| | 3 | | | | | | |
| Noise | 0 | | | | | | |
| | 1 | | | | | | |
| | 2 | | | | | | |
| | 3 | | | | | | |
| **Best** | | 1 | 2 | 0/1 | 1 | 1 | 1/3 |

For the raw-layer method, level 1 was best in the majority of comparisons. Level 0 left dense repeating high-frequency artifacts; level 1 reduced them while preserving structure; levels 2 and 3 over-transformed and washed structure out. On `ch19` with noise, level 1 produced clear hills and trees while level 3 collapsed to abstract colour with no recognisable structure.

Table 18: Transformation-robustness sweep, SAE method.

| Noise | Level | ch10 | ch14 | ch15 | ch21 | ch23 | ch27 |
|---|---|---|---|---|---|---|---|
| No Noise | 0 | | | | | | |
| | 1 | | | | | | |
| | 2 | | | | | | |
| | 3 | | | | | | |

| Noise | Level | ch10 | ch14 | ch15 | ch21 | ch23 | ch27 |
|---|---|---|---|---|---|---|---|
| Noise | 0 | | | | | | |
| | 1 | | | | | | |
| | 2 | | | | | | |
| | 3 | | | | | | |
| **Best** | | Draw | 1/3 | 0/1 | 2 | Draw | 0/3 |

For the SAE method, the picture was more mixed: some channels preferred level 0 or 1, others 2 or 3. The most striking observation was on `ch14` in both noise conditions: strong transformation robustness completely changed the visualization from a muddy crowd-like image into a clear cable-like structure consistent with dataset examples (Section F). We selected level 1 as the default for both methods - aligned with prior literature on small transformation regimes - while noting that the `ch14` behaviour shows transformation robustness can also expose alternative facets of a feature.

### G.8 Total variation penalty

We compared four penalty levels: 0 (disabled), 1 (weight `0.5`), 2 (`1`), and 3 (`5`).

Table 19: Total variation weight sweep, raw-layer method.

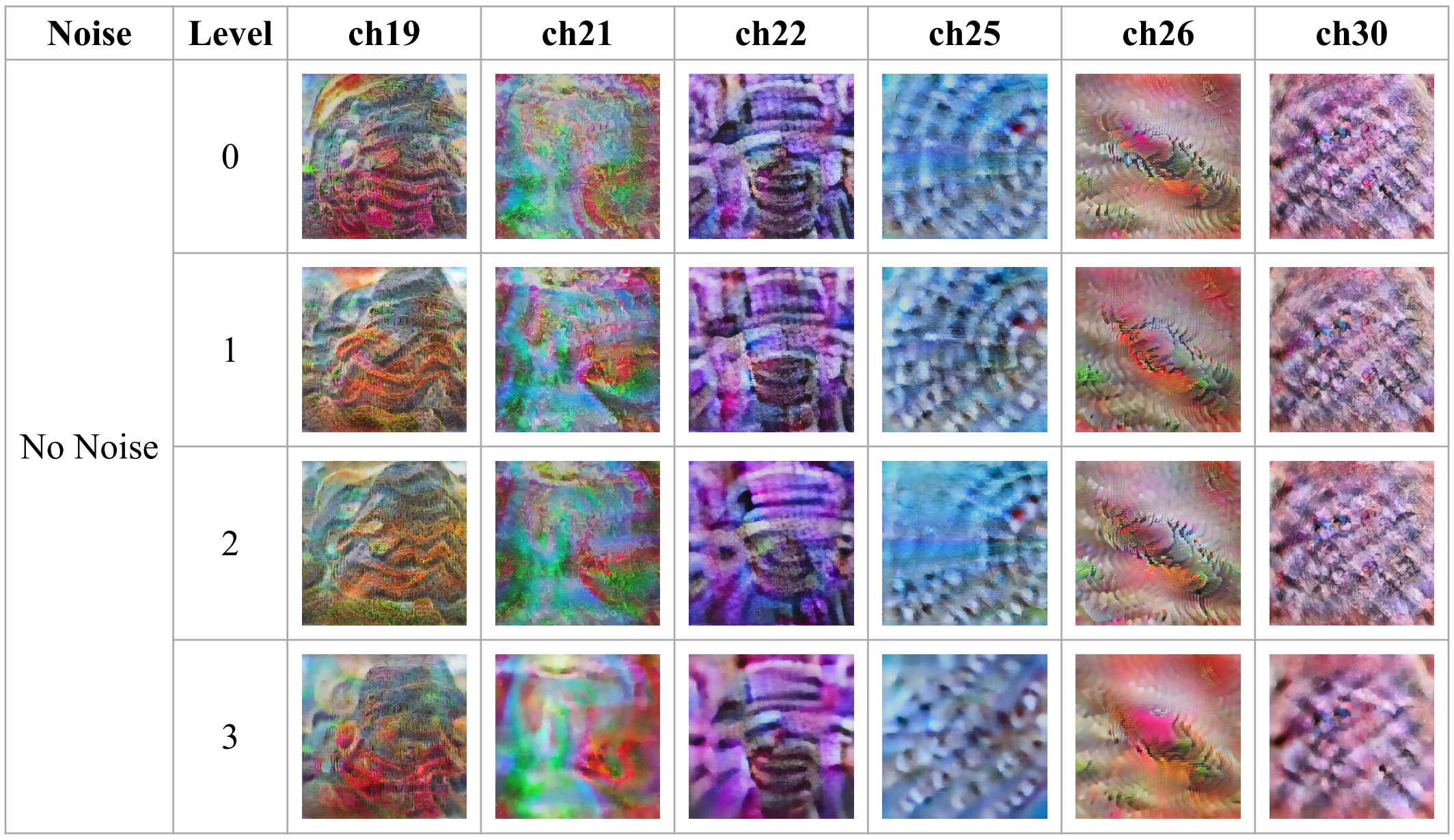

| Noise | Level | ch19 | ch21 | ch22 | ch25 | ch26 | ch30 |
|---|---|---|---|---|---|---|---|
| No Noise | 0 | | | | | | |
| | 1 | | | | | | |
| | 2 | | | | | | |
| | 3 | | | | | | |

| Noise | Level | ch19 | ch21 | ch22 | ch25 | ch26 | ch30 |
|---|---|---|---|---|---|---|---|
| Noise | 0 | | | | | | |
| | 1 | | | | | | |
| | 2 | | | | | | |
| | 3 | | | | | | |
| **Best** | | 1 | 1 | 1 | 0/1 | 0 | 1 |

For the raw-layer method, level 1 (weight `0.5`) was best in the majority of comparisons. Level 0 sometimes left excessive high-frequency noise and fragmented textures; level 1 reduced these artifacts while preserving the underlying structure and colour; levels 2 and 3 over-smoothed. `ch19`, `ch21`, `ch22`, `ch25`, and `ch30` all showed cleaner results at level 1. The exception was `ch26`, where level 0 was best because any TV strength smoothed away the fine feather-like textures characteristic of that channel.

Table 20: Total variation weight sweep, SAE method.

| Noise | Level | ch10 | ch14 | ch15 | ch21 | ch23 | ch27 |
|---|---|---|---|---|---|---|---|
| No Noise | 0 | | | | | | |
| | 1 | | | | | | |
| | 2 | | | | | | |
| | 3 | | | | | | |
| Noise | 0 | | | | | | |

| Noise | Level | ch10 | ch14 | ch15 | ch21 | ch23 | ch27 |
|---|---|---|---|---|---|---|---|
| | 1 | | | | | | |
| | 2 | | | | | | |
| | 3 | | | | | | |
| **Best** | | 0 | 0 | Draw | 0 | 1 | 0 |

For the SAE method, level 0 was best in the majority of comparisons. `ch10`, `ch14`, `ch21`, and `ch27` all produced the most coherent results without any TV penalty; even level 1 began to wash out structure, and levels 2-3 produced nearly uniform images. The exception was `ch23`, where level 1 produced clearer patterns. The SAE method's hyper-sensitivity to total variation prompted us to disable it for the SAE variant. Final weights: `0.5` for the raw-layer method, `0` for the SAE method.

### G.9 Range penalty

We compared four range-penalty levels: 0 (disabled), 1 (weight `0.5`), 2 (`1`), and 3 (`5`).

Table 21: Range penalty weight sweep, raw-layer method.

| Noise | Level | ch19 | ch21 | ch22 | ch25 | ch26 | ch30 |
|---|---|---|---|---|---|---|---|
| No Noise | 0 | | | | | | |
| | 1 | | | | | | |
| | 2 | | | | | | |
| | 3 | | | | | | |
| Noise | 0 | | | | | | |

| Noise | Level | ch19 | ch21 | ch22 | ch25 | ch26 | ch30 |
|---|---|---|---|---|---|---|---|
| | 1 | | | | | | |
| | 2 | | | | | | |
| | 3 | | | | | | |
| **Best** | | 0/1 | 1 | 1 | 0 | 0/Draw | 1 |

For the raw-layer method, level 1 was best in the majority of comparisons, with subtle but consistent improvements: `ch21`, `ch22`, and `ch30` showed better colour balance in both noise conditions, with reduced oversaturation and fewer extreme highlights. `ch19` preferred level 1 only under noise. `ch25` and `ch26` preferred level 0, suggesting that some features benefit from the full dynamic range. Higher levels (2-3) progressively washed colours toward grey.

Table 22: Range penalty weight sweep, SAE method.

| Noise | Level | ch10 | ch14 | ch15 | ch21 | ch23 | ch27 |
|---|---|---|---|---|---|---|---|
| No Noise | 0 | | | | | | |
| | 1 | | | | | | |
| | 2 | | | | | | |
| | 3 | | | | | | |
| Noise | 0 | | | | | | |
| | 1 | | | | | | |

| Noise | Level | ch10 | ch14 | ch15 | ch21 | ch23 | ch27 |
|---|---|---|---|---|---|---|---|
| | 2 | | | | | | |
| | 3 | | | | | | |
| **Best** | | Draw/2 | 1 | 1/Draw | 1 | 1 | 0/1 |

The SAE method tolerated the range penalty well; level 1 was best or equivalent across most channels. `ch14` exhibited the same cable-like transformation observed under transformation robustness: at level 0, a muddy crowd-like texture; at level 1, clear diagonal cable structures. On `ch10` under noise, level 2 produced colourful abstract patterns and level 3 produced a centred reddish-brown mass surrounded by lighter tones, suggesting that the range penalty can also push the optimization toward different interpretations of the same feature. Final weights: `0.5` for both methods.

### G.10 Moment penalty

We compared four moment-penalty levels: 0 (disabled), 1 (weight `0.5`), 2 (`1`), and 3 (`5`).

Table 23: Moment penalty weight sweep, raw-layer method.

| Noise | Level | ch19 | ch21 | ch22 | ch25 | ch26 | ch30 |
|---|---|---|---|---|---|---|---|
| No Noise | 0 | | | | | | |
| | 1 | | | | | | |
| | 2 | | | | | | |
| | 3 | | | | | | |
| Noise | 0 | | | | | | |
| | 1 | | | | | | |

| Noise | Level | ch19 | ch21 | ch22 | ch25 | ch26 | ch30 |
|---|---|---|---|---|---|---|---|
| | 2 | | | | | | |
| | 3 | | | | | | |
| **Best** | | 0/1 | 1/Draw | 0/2 | Draw/1 | 1/Draw | 0/3 |

For the raw-layer method the results were inconsistent across channels and noise conditions: `ch19` preferred level 0 without noise but level 1 with noise; `ch22` preferred level 0 without noise but level 2 with noise; `ch30` preferred level 0 without noise but level 3 with noise. The moment penalty therefore did not provide a reliable improvement for the raw-layer method.

Table 24: Moment penalty weight sweep, SAE method.

| Noise | Level | ch10 | ch14 | ch15 | ch21 | ch23 | ch27 |
|---|---|---|---|---|---|---|---|
| No Noise | 0 | | | | | | |
| | 1 | | | | | | |
| | 2 | | | | | | |
| | 3 | | | | | | |
| Noise | 0 | | | | | | |
| | 1 | | | | | | |
| | 2 | | | | | | |

| Noise | Level | ch10 | ch14 | ch15 | ch21 | ch23 | ch27 |
|---|---|---|---|---|---|---|---|
| | 3 | | | | | | |
| **Best** | | 0 | 1 | 1 | 2 | 0 | 1/2 |

The SAE method was more consistent. `ch14` exhibited the cable-like transformation in **both** noise conditions at level 1 (under range penalty the transformation only appeared without noise). `ch21` preferred level 2 in both conditions; `ch15` showed slightly more defined diagonal strokes at level 1. The moment penalty's clearest contribution was helping the optimizer escape uninterpretable local maxima, as the `ch14` case illustrates. Final weights: `0` for the raw-layer method, `0.5` for the SAE method.

### G.11 Gradient smoothing

We compared four smoothing levels: 0 (disabled), 1 (sigma `0.5`), 2 (`1`), and 3 (`5`).

Table 25: Gradient-smoothing sweep, raw-layer method.

| Noise | Level | ch19 | ch21 | ch22 | ch25 | ch26 | ch30 |
|---|---|---|---|---|---|---|---|
| No Noise | 0 | | | | | | |
| | 1 | | | | | | |
| | 2 | | | | | | |
| | 3 | | | | | | |
| Noise | 0 | | | | | | |
| | 1 | | | | | | |
| | 2 | | | | | | |

| Noise | Level | ch19 | ch21 | ch22 | ch25 | ch26 | ch30 |
|---|---|---|---|---|---|---|---|
| | 3 | | | | | | |
| **Best** | | 1 | 1 | 1 | 1 | 1/Draw | 0/1 |

For the raw-layer method, level 1 was best in the majority of comparisons, producing more defined shapes and increased contrast on `ch19`, `ch21`, `ch22`, `ch25`, and `ch26`. `ch30` preferred level 0 without noise but level 1 under noise. Levels 2-3 over-blurred, although the effect was mild.

Table 26: Gradient-smoothing sweep, SAE method.

| Noise | Level | ch10 | ch14 | ch15 | ch21 | ch23 | ch27 |
|---|---|---|---|---|---|---|---|
| No Noise | 0 | | | | | | |
| | 1 | | | | | | |
| | 2 | | | | | | |
| | 3 | | | | | | |
| Noise | 0 | | | | | | |
| | 1 | | | | | | |
| | 2 | | | | | | |
| | 3 | | | | | | |
| **Best** | | Draw | 0 | 0 | 0 | Draw | 0 |

The SAE method strongly preferred no smoothing. On `ch14`, level 0 preserved the vivid cable-like diagonal structures that any smoothing quickly blurred; `ch15`, `ch21`, and `ch27` similarly degraded as smoothing increased. `ch10` and `ch23` showed no significant difference across levels. Final values: `0.5` for the raw-layer method, `0` for the SAE method.

## NeurIPS Paper Checklist

1. **Claims**

   Question: Do the main claims made in the abstract and introduction accurately reflect the paper's contributions and scope?

   Answer: [Yes]

   Justification: The abstract and Section 1 list the four contributions (proposing LVO, raw-layer baseline, calibration study, complementary insight) and the empirical scope (Stable Diffusion 1.5 fine-tuned on Style50, layer `up1.1`, qualitative evaluation). All four are addressed in Sections 2–6, and Section 7 explicitly bounds the scope.

2. **Limitations**

   Question: Does the paper discuss the limitations of the work performed by the authors?

   Answer: [Yes]

   Justification: Section 7 (Limitations) discusses the qualitative-only evaluation, the single-model / single-layer / 30-channel scope, the possibility of multifaceted features, and the dependence on SAE quality.

3. **Theory assumptions and proofs**

   Question: For each theoretical result, does the paper provide the full set of assumptions and a complete (and correct) proof?

   Answer: [NA]

   Justification: The paper does not contain theoretical results.

4. **Experimental result reproducibility**

   Question: Does the paper fully disclose all the information needed to reproduce the main experimental results?

   Answer: [Yes]

   Justification: Section 2 specifies the algorithm; Section 3 specifies the model, layer, dataset, evaluation protocol, optimizer, learning rate, number of steps, and hardware; Section 4 reports the final hyperparameters in Table 1. Code and configuration are released as supplementary material.

5. **Open access to data and code**

   Question: Does the paper provide open access to the data and code, with sufficient instructions to faithfully reproduce the main experimental results?

   Answer: [Yes]

   Justification: Source code, configuration files, and reproduction instructions are released at `github.com/aszokalski/diffusion-deep-dream-research`. All datasets and SAE checkpoints used are obtained from the original public releases cited in Section 3.

6. **Experimental setting/details**

   Question: Does the paper specify all the training and test details necessary to understand the results?

   Answer: [Yes]

   Justification: Section 3 reports the model, layer, dataset, evaluation protocol, optimizer (Adam), learning rate (`0.05`), number of steps (`100`), seed handling, and compute (single A100 GPU).

Section 4 reports the calibration procedure and Table 1 reports the final hyperparameters for both methods.

7. **Experiment statistical significance**

   Question: Does the paper report error bars suitably and correctly defined or other appropriate information about the statistical significance of the experiments?

   Answer: [No]

   Justification: Evaluation is qualitative. We instead report consistency across multiple activity peaks and across random seeds for each feature (Sections 5.2.1, 5.2.2 and the appendix), which is the form of reproducibility appropriate to feature visualizations [3], [4], [10], [12], [15].

8. **Experiments compute resources**

   Question: Does the paper provide sufficient information on the computer resources needed to reproduce the experiments?

   Answer: [Yes]

   Justification: Section 3 reports the per-visualization cost ( `1` minute on a single A100 GPU at `100` Adam steps, learning rate `0.05`) and notes that the full study, including hyperparameter sweeps, was run on the PLGrid Cyfronet AGH cluster. SAE training is reused from prior work and was not performed here.

9. **Code of ethics**

   Question: Does the research conducted in the paper conform, in every respect, with the NeurIPS Code of Ethics?

   Answer: [Yes]

   Justification: The work uses publicly released models and datasets, introduces no human-subject component, and releases no new generative capability. See also Section 8.

10. **Broader impacts**

    Question: Does the paper discuss both potential positive societal impacts and negative societal impacts of the work performed?

    Answer: [Yes]

    Justification: Section 8 discusses the defensive profile (auditing, unlearning, content moderation) and acknowledges the dual-use risk that feature-level interpretability could in principle help locate features useful for evading filters; we assess this risk as low because no new generative capability or attack vector is introduced.

11. **Safeguards**

    Question: Does the paper describe safeguards for responsible release of data or models that have a high risk for misuse?

    Answer: [NA]

    Justification: No new pretrained generators, datasets, or other high-risk assets are released. The visualization framework operates on the publicly released SAeUron SAEs and a Style50 fine-tune of Stable Diffusion 1.5 obtained from their original releases.

12. **Licenses for existing assets**

    Question: Are the creators or original owners of assets used in the paper properly credited and are the license and terms of use explicitly mentioned and properly respected?

    Answer: [Yes]

Justification: The base Stable Diffusion 1.5 weights are released under the CreativeML Open-RAIL-M license; the UnlearnCanvas dataset and Style50 fine-tune are released under the MIT license; and the SAeUron code and SAE checkpoints are released under the Apache-2.0 license. All assets are cited at the point of use (Section 3) and used under their original licenses. We do not redistribute any of these assets.

13. **New assets**

    Question: Are new assets introduced in the paper well documented and is the documentation provided alongside the assets?

    Answer: [Yes]

    Justification: We release the experiment framework as source code at `github.com/aszokalski/diffusion-deep-dream-research` under a BSD 3-Clause-style license, with a README, a Hydra configuration directory documenting every hyperparameter, and reproduction scripts for each pipeline stage. No new datasets or pretrained models are released.

14. **Crowdsourcing and research with human subjects**

    Question: For crowdsourcing experiments and research with human subjects, does the paper include the full text of instructions given to participants, screenshots, and details about compensation?

    Answer: [NA]

    Justification: The paper does not involve crowdsourcing or research with human subjects.

15. **Institutional review board (IRB) approvals**

    Question: Does the paper describe potential risks incurred by study participants and whether IRB approvals were obtained?

    Answer: [NA]

    Justification: The paper does not involve research with human subjects.

16. **Declaration of LLM usage**

    Question: Does the paper describe the usage of LLMs if it is an important, original, or non-standard component of the core methods in this research?

    Answer: [NA]

    Justification: No LLM is part of the core methodology. Any LLM use was limited to writing assistance and does not affect scientific rigor or originality.